\documentclass[12pt]{article}
\usepackage{amsmath}
\usepackage{times}
\usepackage{graphicx}
\usepackage{color}
\usepackage{multirow}
\usepackage[natbibapa]{apacite} 
\usepackage{rotating}
\usepackage{bbm}
\usepackage{latexsym}
\usepackage{hyperref}
\usepackage{url}
\usepackage{amsfonts}
\usepackage{xcolor}
\usepackage[section]{placeins}
\usepackage{graphicx}
\usepackage{multirow} 
\usepackage{booktabs,floatrow}
\usepackage{microtype}

\usepackage[english]{babel}

\newcommand\thl[1]{{#1}}

\newcommand{\captionfonts}{\normalsize}

\makeatletter  
\long\def\@makecaption#1#2{%
  \vskip\abovecaptionskip
  \sbox\@tempboxa{{\captionfonts #1: #2}}%
  \ifdim \wd\@tempboxa >\hsize
    {\captionfonts #1: #2\par}
  \else
    \hbox to\hsize{\hfil\box\@tempboxa\hfil}%
  \fi
  \vskip\belowcaptionskip}
\makeatother   

\begin{document}
\hspace{13.9cm}1

\ \vspace{20mm}\\

{\LARGE Parametric UMAP embeddings for representation and semi-supervised learning}

\ \\
{\bf \large Tim Sainburg$^{\displaystyle 1}$, Leland McInnes$^{\displaystyle 2}$, Timothy Q Gentner$^{\displaystyle 1}$}\\
{$^{\displaystyle 1}$UC San Diego}\\
{$^{\displaystyle 2}$Tutte Institute for Mathematics and Computing
}\\
%


\thispagestyle{empty}
\markboth{}{NC instructions}
\ \vspace{-0mm}\\
%
\begin{center} {\bf Abstract} \end{center}
UMAP is a non-parametric graph-based dimensionality reduction algorithm using applied Riemannian geometry and algebraic topology to find low-dimensional embeddings of structured data. The UMAP algorithm consists of two steps: (1) Compute a graphical representation of a dataset (fuzzy simplicial complex), and (2) Through stochastic gradient descent, optimize a low-dimensional embedding of the graph. 
Here, we extend the second step of UMAP to a parametric optimization over neural network weights, learning a parametric relationship between data and embedding.
We first demonstrate that Parametric UMAP performs comparably to its non-parametric counterpart while conferring the benefit of a learned parametric mapping (e.g. fast online embeddings for new data). We then explore UMAP as a regularization, constraining the latent distribution of autoencoders, parametrically varying global structure preservation, and improving classifier accuracy for semi-supervised learning by capturing structure in unlabeled data.\footnote{\href{https://colab.research.google.com/drive/1WkXVZ5pnMrm17m0YgmtoNjM_XHdnE5Vp?usp=sharing}{\textcolor{blue}{Google Colab walkthrough}}}

\section{Introduction}

Current non-linear dimensionality reduction algorithms can be divided broadly into non-parametric algorithms which rely on the efficient computation of probabilistic relationships from neighborhood graphs to extract structure in large datasets (e.g. UMAP \citep{mcinnes2018umap}, t-SNE \citep{van2008visualizing}, LargeVis \citep{tang2016visualizing}), and parametric algorithms, which, driven by advances in deep-learning, optimize an objective function related to capturing structure in a dataset over neural network weights (e.g. \citealt{hinton2006reducing, ding2018interpretable, ding2019deep, szubert2019structure, kingma2013auto}). 

In recent years, a number of parametric dimensionality reduction algorithms have been developed to wed these two classes of methods, learning a structured graphical representation of the data and using a deep neural network to capture that structure (discussed in \ref{sec:related_works}). In particular, over the past decade, several variants of the t-SNE algorithm have proposed parameterized forms of t-SNE \citep{van2009learning, gisbrecht2015parametric, bunte2012general, gisbrecht2012out}. Parametric t-SNE \citep{van2009learning} for example, trains a deep neural network to minimize loss over a t-SNE graph. However, the t-SNE loss function itself is not well-suited to neural network training paradigms. In particular, t-SNE's optimization requires normalization over the entire dataset at each step of optimization, making batch-based optimization and online learning of large datasets difficult. 
In contrast, UMAP is optimized using negative sampling \citep{mikolov2013distributed, tang2016visualizing} to sparsely sample edges during optimization, making it, in principle, more well-suited to batch-wise training as is common deep learning applications. 
Our proposed method, Parametric UMAP, brings the non-parametric graph-based dimensionality reduction algorithm UMAP into an emerging class of parametric topologically-inspired embedding algorithms. 

In the following section, we broadly outline the algorithm underlying UMAP to explain why our proposed algorithm, Parametric UMAP, is particularly well suited to deep learning applications. We contextualize our discussion of UMAP in t-SNE, to outline the advantages that UMAP confers over t-SNE in the domain of parametric neural-network-based embedding. We then perform experiments comparing our algorithm, Parametric UMAP, to parametric and non-parametric algorithms. Finally, we show a novel extension of Parametric UMAP to semi-supervised learning.

\begin{figure*}[!htbp]
  \centering
      \includegraphics[width=1\textwidth]{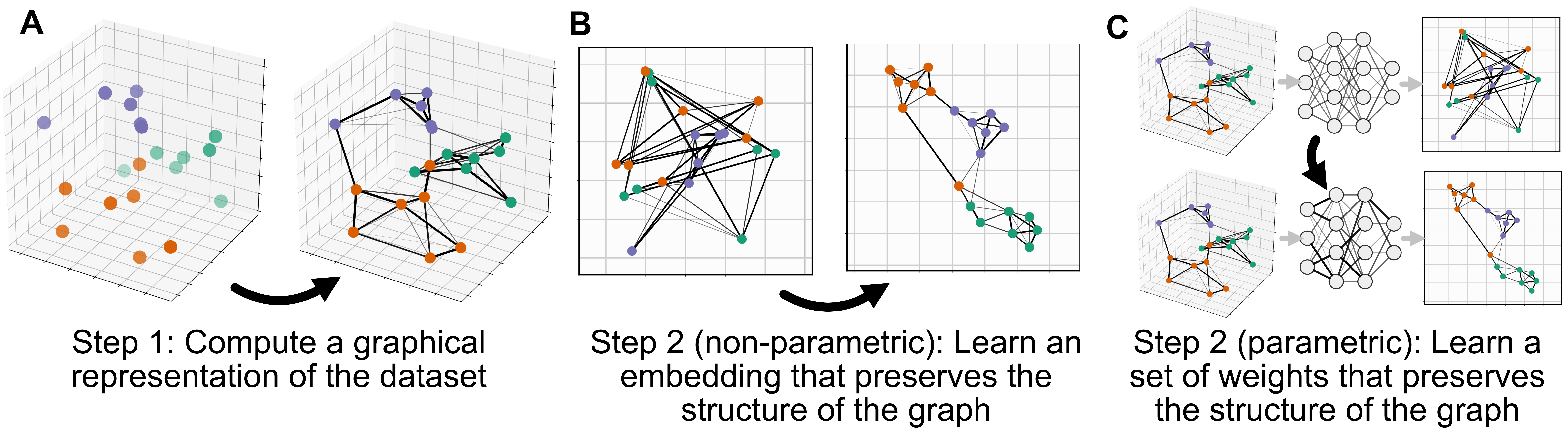}
  \caption{Overview of UMAP (A $\rightarrow$ B) and Parametric UMAP (A $\rightarrow$ C). 
  }
\label{fig:overview}
\end{figure*}
\section{Parametric and non-parametric UMAP}

\label{sec:umap_tsne}
UMAP and t-SNE have the same goal: Given a $D$-dimensional data set $\mathbf{X} \in \mathbb{R}^{D}$, produce a $d$ dimensional embedding $Z \in \mathbb{R}^d$ such that points that are close together in $X$ (e.g. $x_i$ and $x_j$) are also close together in $Z$ ($z_i$ and $z_j$).

Both algorithms are comprised of the same two broad steps: first construct a graph of local relationships between datasets (Fig \ref{fig:overview}A), then optimize an embedding in low dimensional space which preserves the structure of the graph (Fig \ref{fig:overview}B). The parametric approach replaces the second step of this process with an optimization of the parameters of a deep neural network over batches (Fig \ref{fig:overview}C). 
To understand how Parametric UMAP is optimized, it is necessary to understand these steps.

\subsection{Graph Construction}

\paragraph{Computing probabilities in $X$}

The first step in both UMAP and t-SNE is to compute a distribution of probabilities $P$ between pairs of points in $X$ based upon the distances between points in data space. Probabilities are initially computed as local, one-directional, probabilities between a point and its neighbors in data-space, then symmetrized to yield a final probability representing the relationship between pairs of points.

In t-SNE, these probabilities are treated as conditional probabilities of neighborhood ($p^{\textrm{\tiny{t-SNE}}}_{i|j}$) computed using a Gaussian distribution centered at $x_i$.

\begin{equation}
p_{j \mid i}^{\textrm{\tiny{t-SNE}}}=\frac{\exp \left(-\textrm{d}( \mathbf{x}_{i},\mathbf{x}_{j}) / 2 \sigma_{i}^{2}\right)}{\sum_{k \neq i} \exp \left(-\textrm{d}( \mathbf{x}_{i},\mathbf{x}_{k}) / 2 \sigma_{i}^{2}\right)}
\label{conditional_prob_tsne}
\end{equation}

Where $\textrm{d}( \mathbf{x}_{i},\mathbf{x}_{j})$ represents the distance between ${x}_{i}$ an ${x}_{j}$ (e.g. Euclidean distance) and $\sigma_{i}$ is the standard deviation for the Gaussian distribution, set based upon the a perplexity parameter such that one standard deviation of the Gaussian kernel fits a a set number of nearest-neighbors in $X$.  

In UMAP, local, one-directional, probabilities ($P^{\textrm{\tiny{UMAP}}}_{i|j}$) are computed between a point and its neighbors to determine the probability with which an edge (or simplex) exists, based upon an assumption that data is uniformly distributed across a manifold in a warped dataspace. Under this assumption, a local notion of distance is set by the distance to the $k$\textsuperscript{th} nearest neighbor and the local probability is scaled by that local notion of distance.
\begin{equation}
p_{j \mid i}^{\textrm{\tiny{UMAP}}} = \exp( -(\textrm{d}( \mathbf{x}_{i},\mathbf{x}_{j}) - \rho_{i}) / \sigma_{i})
\label{conditional_prob_umap}
\end{equation}

Where $\rho_{i}$ is a local connectivity parameter set to the distance from $x_i$ to its nearest neighbor, and $\sigma_{i}$ is a local connectivity parameter set to match the local distance around $x_i$ upon its $k$ nearest neighbors (where $k$ is a hyperparameter). 

After computing the one-directional edge probabilities for each datapoint, UMAP computes a global probability as the probability of either of the two local, one-directional, probabilities occurring:  
\begin{equation}
p_{i j}^{\textrm{\tiny{UMAP}}}=\left(p_{j \mid i}+p_{i \mid j}\right)-p_{j \mid i} p_{i \mid j}
\label{global_probability_umap}
\end{equation}

In contrast, t-SNE symmetrizes the conditional probabilities as

\begin{equation}
p_{i j}^{\textrm{\tiny{t-SNE}}}=\frac{p_{j \mid i}+p_{i \mid j}}{2 N}
\label{global_probability_tsne}
\end{equation}

\subsection{Graph Embedding}

After constructing a distribution of probabilistically weighted edges between points in $X$, UMAP and t-SNE initialize an embedding in $Z$ corresponding to each data point, where a probability distribution ($Q$) is computed between points as was done with the distribution ($P$) in the input space. The objective of UMAP and t-SNE is then to optimize that embedding to minimize the difference between $P$ and $Q$.

\paragraph{Computing probabilities in $Z$}

In embedding space, the pairwise probabilities are computed directly without first computing local, one-directional probabilities. 

In the t-SNE embedding space, the pairwise probability between two points $q^{\textrm{\tiny{t-SNE}}}_{i|j}$ is computed in a similar manner to  $p^{\textrm{\tiny{t-SNE}}}_{i|j}$, but where the Gaussian distribution is replaced with the fatter-tailed Student's t-distribution (with one degree of freedom), which is used to overcome the 'crowding problem' \citep{van2008visualizing} in translating volume differences in high-dimensional spaces to low-dimensional spaces: 

\begin{equation}
q_{i j}^{\textrm{\tiny{t-SNE}}}=\frac{\left(1+\left\|\mathbf{z}_{i}-\mathbf{z}_{j}\right\|^{2}\right)^{-1}}{\sum_{k \neq l}\left(1+\left\|\mathbf{z}_{k}-\mathbf{z}_{l}\right\|^{2}\right)^{-1}}
\label{prob_tsne_q}
\end{equation}

UMAP's computation of the pairwise probability $q^{\textrm{\tiny{UMAP}}}_{ij}$ between points in the embedding space $Z$ uses a different family of functions: 

\begin{equation}
q_{i j}^{\textrm{\tiny{UMAP}}}=\left(1+a\ \|z_{i}-z_{j}||^{2 b}\right)^{-1}
\label{prob_umap_q}
\end{equation}

Where $a$ and $b$ are hyperparameters set based upon a desired minimum distance between points in embedding space. Notably, the UMAP probability distribution in embedding space is not normalized, while the t-SNE distribution is normalized across the entire distribution of probabilities, meaning that the entire distribution of probabilities needs to be calculated before each optimization step of t-SNE.

\paragraph{Cost function}

Finally, the distribution of embeddings in $Z$ is optimized to minimize the difference between $Q$ and $P$. 

In t-SNE, a Kullback-Leibler divergence between the two probability distributions is used, and gradient descent in t-SNE is computed over the embeddings:

\begin{equation}
C_{\textrm{\tiny{t-SNE}}}=\sum_{i \neq j} p_{i j} \log \frac{p_{i j}}{q_{i j}} 
\label{tsne_cost}
\end{equation}

In UMAP, the cost function is cross-entropy, also optimized using gradient descent:

\begin{equation}
C_{\textrm{\tiny{UMAP}}}= \sum_{i \neq j} p_{i j} \log \left(\frac{p_{i j}}{q_{i j}}\right)+\left(1-p_{i j}\right) \log \left(\frac{1-p_{i j}}{1-q_{i j}}\right)
\label{umap_cost}
\end{equation}

\subsection{Attraction and repulsion}


Minimizing the cost function over every possible pair of points in the dataset would be computationally expensive. UMAP and more recent variants of t-SNE both use shortcuts to bypass much of that computation. In UMAP, those shortcuts are directly advantageous to batch-wise training in a neural network.

The primary intuition behind these shortcuts is that the cost function of both t-SNE and UMAP can both be broken out into a mixture of attractive forces between locally connected embeddings and repulsive forces between non-locally connected embeddings.

\paragraph{Attractive forces}
Both UMAP and t-SNE utilize a similar strategy in minimizing the computational cost over attractive forces: they rely on an approximate nearest neighbors graph\footnote{UMAP requires substantially fewer nearest neighbors than t-SNE, which generally requires 3 times the perplexity hyperparameter (defaulted at 30 here), whereas UMAP computes only 15 neighbors by default, which is computationally less costly.}. The intuition for this approach is that elements that are further apart in data space have very small edge probabilities, which can be treated as effectively zero. Thus, edge probabilities and attractive forces only need to be computed over the nearest neighbors, non-nearest neighbors can be treated as having an edge probability of zero. Because nearest-neighbor graphs are themselves computationally expensive, approximate nearest neighbors (e.g. \citealt{dong2011efficient}) produce effectively similar results. 

\paragraph{Repulsive forces}
Because most datapoints are not locally connected, we do not need to waste computation on most pairs of embeddings. 

UMAP takes a shortcut motivated by the language model word2vec \citep{mikolov2013distributed} and performs negative sampling over embeddings. 
Each training step iterates over positive, locally connected, edges and randomly samples edges from the remainder of the dataset treating their edge probabilities as zero to compute cross-entropy. Because most datapoints are not locally connected and have a very low edge probability, these negative samples are, on average, correct, allowing UMAP to sample only sparsely over edges in the dataset. 

In t-SNE, repulsion is derived from the normalization of $Q$. A few methods for minimizing the amount of computation needed for repulsion have been developed. The first is the Barnes-Hut tree algorithm \citep{van2014accelerating}, which bins the embedding space into cells and where repulsive forces can be computed over cells rather than individual datapoints within those cells. Similarly, the more recent interpolation-based t-SNE (FIt-SNE; \citealt{linderman2017efficient,linderman2019fast}) divides the embedding space up into a grid and computes repulsive forces over the grid, rather than the full set of embeddings.

\subsection{Parametric UMAP}

To summarize, both t-SNE and UMAP rely on the construction of a graph, and a subsequent embedding that preserves the structure of that graph (Fig. \ref{fig:overview}). UMAP learns an embedding by minimizing cross-entropy sampled over positively weighted edges (attraction) and using negative sampling randomly over the dataset (repulsion), allowing minimization to occur over sampled batches of the dataset. t-SNE, meanwhile, minimizes a KL divergence loss function normalized over the entire set of embeddings in the dataset using different approximation techniques to compute attractive and repulsive forces. 

Because t-SNE optimization requires normalization over the distribution of embedding in projection space, gradient descent can only be performed after computing edge probabilities over the entire dataset. Projecting an entire dataset into a neural network between each gradient descent step would be too computationally expensive to optimize, however. 
The trick that Parametric t-SNE proposes for this problem is to split the dataset up into large batches (e.g. 5000 data points in the original paper)
that are used to compute separate graphs
that are independently normalized over and used constantly throughout training, meaning that relationships between elements in different batches are not explicitly preserved. Conversely, a parametric form of UMAP, by using negative sampling, can in principle be trained on batch sizes as small as a single edge, making it suitable for minibatch training needed for memory-expensive neural networks trained on the full graph over large datasets as well as online learning.

Given these design features, UMAP loss can be applied as a regularization in typical stochastic gradient descent deep learning paradigms, without requiring the batching trick that Parametric T-SNE relies upon. Despite this, a parametric extension to the UMAP learning algorithm has not yet been explored. Here, we explore the performance of a parametric extension to UMAP relative to current embedding algorithms and perform several experiments further extending Parametric UMAP to novel applications \footnote{See code implementations: Experiments \url{https://github.com/timsainb/ParametricUMAP_paper} Python package \url{https://github.com/lmcinnes/umap}}.

\section{Related Work}
\label{sec:related_works}
Beyond Parametric t-SNE and Parametric UMAP, a number of recent parametric dimensionality reduction algorithms utilizing structure-preserving constraints exist which were not compared here.
This work is relevant to ours and is mentioned here to provide clarity on the current state of parametric topologically motivated and structure-preserving dimensionality reduction algorithms.

Moor et al., (topological autoencoders; \citeyear{moor2019topological}) and Hoffer et al. (Connectivity-Optimized Representation Learning; \citeyear{hofer2019connectivity}) apply an additional topological structure-preserving loss using persistent homology over mini-batches to the latent space of an autoencoder. 
Jia et al., (Laplacian Autoencoders; \citeyear{jia2015laplacian}) similarly defines an autoencoder with a local structure preserving regularization.
Mishne et al., (Diffusion Nets; \citeyear{mishne2019diffusion}) define an autoencoder extension based upon diffusion maps that constrains the latent space of the autoencoder. 
Ding et al., (scvis; \citeyear{ding2018interpretable}) and Graving and Couzin (VAE-SNE; \citeyear{graving2020vae}) describe VAE-derived dimensionality reduction algorithms based upon the ELBO objective. 
Duque et al (geometry-regularized autoencoders; \citeyear{duque2020extendable}) regularize an autoencoder with the PHATE (Potential of Heat-diffusion for Affinity-based Trajectory Embedding) embedding algorithm \citep{moon2017phate}.
Szubert et al (ivis; \citeyear{szubert2019structure}) and Robinson (Differential Embedding Networks; \citeyear{robinson2020interpretable}) make use of Siamese neural network architectures with structure-preserving loss functions to learn embeddings.
Pai et al., (DIMAL; \citeyear{pai2019dimal}) similarly uses Siamese networks constrained to preserve geodesic distances for dimensionality reduction. Several of these parametric approaches indirectly condition neural networks (e.g. autoencoders) on non-parametric embeddings rather than directly upon the loss of the algorithm, which can be applied to arbitrary embedding algorithms. We contrast indirect and direct parametric embeddings in \ref{sec:MSE}.

\section{UMAP as a regularization}

\begin{figure*}[!htbp]
  \centering
      \includegraphics[width=1\textwidth]{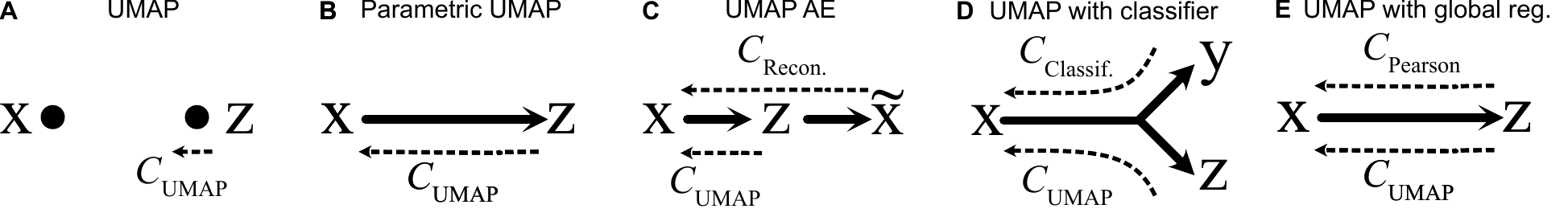}
  \caption{Varients of UMAP used in this paper. Solid lines represent neural networks. Dashed lines represent error gradients.}
\label{fig:networks}
\end{figure*}

In machine learning, regularization refers to the modification of a learning algorithm to improve generalization to new data. Here, we consider both regularizing neural networks with UMAP loss, as well as using additional loss functions to regularize the embedding that UMAP learns. While non-parametric UMAP optimizes UMAP loss directly over embeddings (Fig \ref{fig:networks}A), our proposed algorithm, Parametric UMAP, applies the same cost function over an encoder network (Fig 2B). By applying additional losses, we can use both regularize UMAP with, as well as use UMAP to, regularize additional training objectives, which we outline below.

\subsection{Autoencoding with UMAP}

AEs are by themselves a powerful dimensionality reduction algorithm \citep{hinton2006reducing}. Thus, combining them with UMAP may yield additional benefits in capturing latent structure. We used an autoencoder as an additional regularization to Parametric UMAP (Fig \ref{fig:networks}C). 
A UMAP/AE hybrid is simply the combination of the UMAP loss and a reconstruction loss, both applied over the network. VAEs have similarly been used in conjunction with Parametric t-SNE for capturing structure in animal behavioral data \citep{graving2020vae} and combining t-SNE, which similarly emphasizes local structure, with AEs aids in capturing more global structure over the dataset \citep{van2008visualizing, graving2020vae}. 

\subsection{Semi-supervised learning}

Parametric UMAP can be used to regularize supervised classifier networks, training the network on a combination of labeled data with the classifier loss and unlabeled data with UMAP loss (Fig \ref{fig:networks}D). Semi-supervised learning refers to the use of unlabeled data to jointly learn the structure of a dataset while labeled data is used to optimize the supervised objective function, such as classifying images. Here, we explore how UMAP can be jointly trained as an objective function in a deep neural network alongside a classifier. 

\begin{figure}[!htb]
  \centering
      \includegraphics[width=1\textwidth]{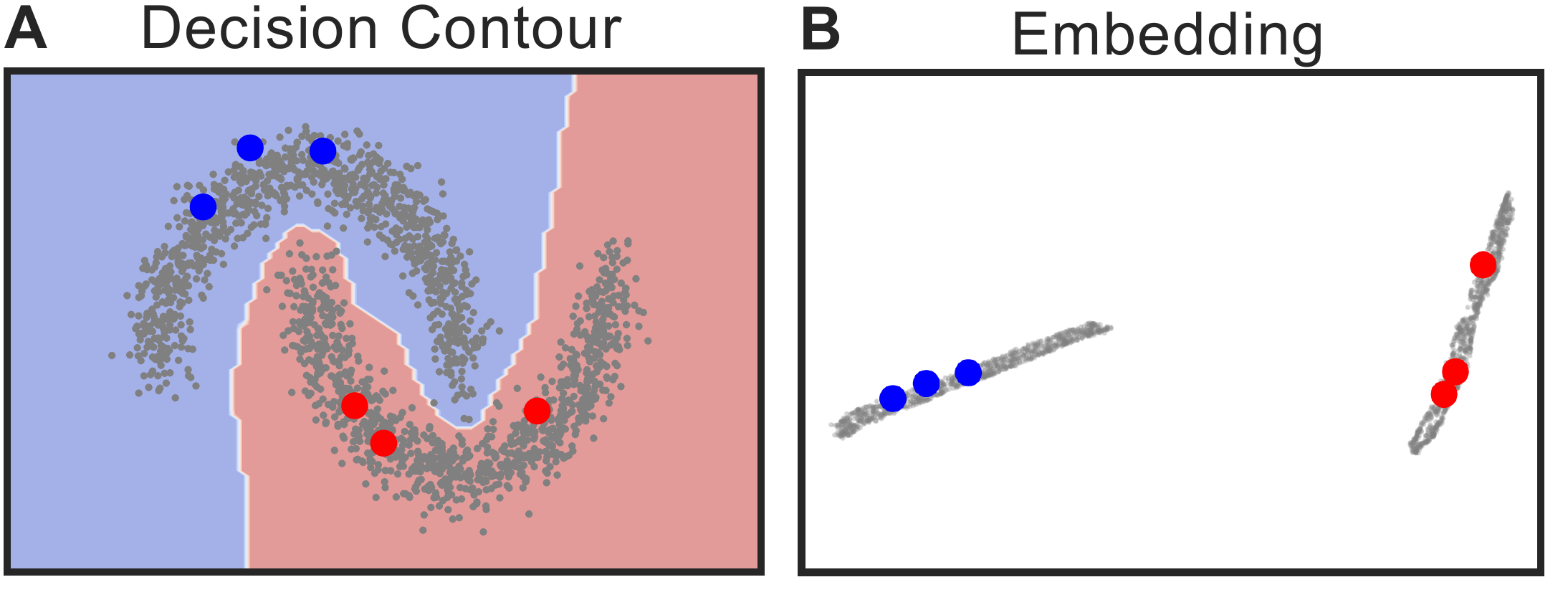}
  \caption{An example of semi-supervised learning with UMAP on the moons dataset. 
  }
\label{fig:moons}

\end{figure}

In the example in Fig \ref{fig:moons}, we show an intuitive example of semi-supervised learning using UMAP over the Moons dataset \citep{pedregosa2011scikit}. By training a Y-shaped network (Fig \ref{fig:networks}D) both on the classifier loss over labeled datapoints (Fig \ref{fig:moons}A, red and blue) and the UMAP loss over unlabeled datapoints (Fig \ref{fig:moons}A, grey) jointly, the shared latent space between the UMAP and classifier network pulls apart the two moons (Fig \ref{fig:moons}B), resulting in a decision boundary that divides cleanly between the two distributions in dataspace. 

\subsection{Preserving global structure}
An open issue in dimensionality reduction is how to balance local and global structure preservation \citep{de2003global, becht2019dimensionality, kobak2021initialization}. Algorithms that rely on sparse nearest neighbor graphs, like UMAP, focus on capturing the local structure present between points and their nearest neighbors, while global algorithms, like Multi-Dimensional Scaling (MDS), attempt to preserve all relationships during embedding. 
Local algorithms are both computationally more efficient and capture structure that is lost in global algorithms (e.g. the clusters corresponding to numbers found when projecting MNIST into UMAP). While local structure preservation captures more application-relevant structure in many datasets, however, the ability to additionally capture global structure is still often desirable. 
The approach used by non-parametric t-SNE and UMAP is to initialize embeddings with global structure-preserving embeddings such as PCA or Laplacian eigenmaps embeddings.
In Parametric UMAP, we explore a different tactic, imposing global structure by jointly training on a global structure preservation loss directly.

\section{Experiments}
\label{sec:experiments}

\begin{figure*}[!htbp]
  \centering
      \includegraphics[width=1.0\textwidth]{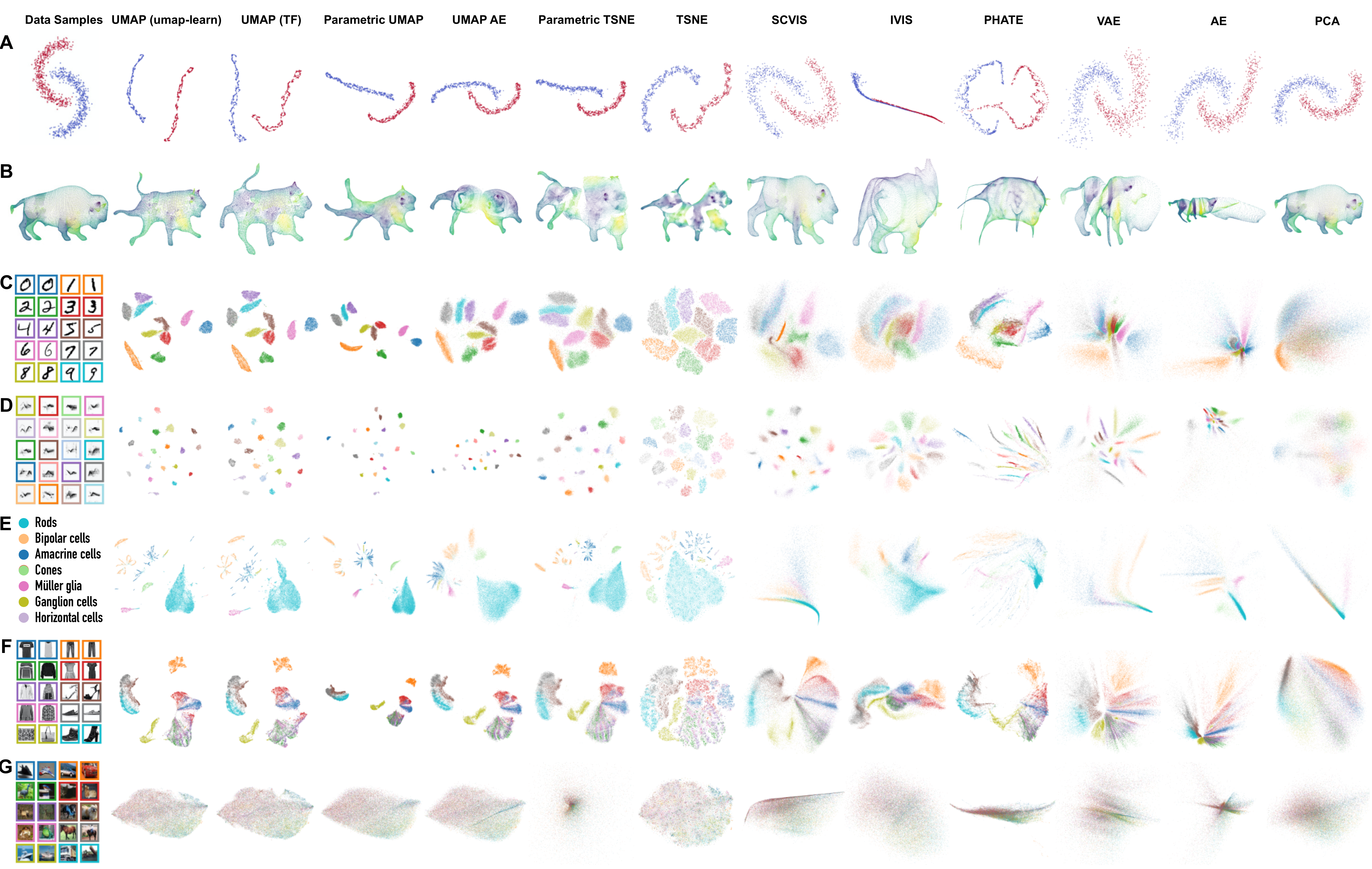}
  \caption{Comparison of projections from multiple datasets using UMAP, UMAP in Tensorflow, Parametric UMAP, Parametric UMAP with an Autoencoder loss, Parametric t-SNE, t-SNE, SCVIS, IVIS, PHATE, a VAE, an AE, and PCA. (a) Moons. (B) 3D buffalo. (c) MNIST (d) Cassin's vireo song segments (e) Mouse retina single-cell transcriptomes. (f) Fashion MNIST (g) CIFAR10. The Cassin's vireo dataset uses a dynamic time warping loss and an LSTM network for the encoder and decoder for the neural networks. The image datasets use a convnet for the encoder and decoder for the neural networks. The bison examples use a t-SNE perplexity of 500 and 150 nearest neighbors in UMAP to capture more global structure.}
  \label{fig:projections}
\end{figure*}

Experiments were performed comparing Parametric UMAP and a UMAP/AE hybrid, to several baselines: nonparametric UMAP, nonparametric t-SNE (FIt-SNE) \citep{linderman2019fast, Poliar731877}, Parametric t-SNE, an AE, a VAE, and PCA projections. As additional baselines, we compared PHATE (non-parametric), SCVIS (parametric), and IVIS (parametric) which are described in the related works section (\ref{sec:related_works}).
We also compare a second non-parametric UMAP implementation that has the same underlying code as Parametric UMAP, but where optimization is performed over embeddings directly, rather than neural network weights. 
This comparison is made to provide a bridge between the UMAP-learn implementation and parametric UMAP, to control for any potential implementation differences. Parametric t-SNE, Parametric UMAP, the AE, VAE, and the UMAP/AE hybrid use the same neural network architectures and optimizers within each dataset (described in \ref{sec:datasets} and \ref{embedding_algorithms}). 

We used the common machine learning benchmark datasets MNIST, FMNIST, and CIFAR10 alongside two real-world datasets in areas where UMAP has proven a useful tool for dimensionality reduction: a single-cell retinal transcriptome dataset \citep{macosko2015highly}, and a bioacoustic dataset of Cassin's vireo song, recorded in the Sierra Nevada mountains \citep{hedley2016complexity, hedley2016composition}.

\subsection{Embeddings}
\label{sec:embeddings}

We first confirm that Parametric UMAP produces embeddings that are of a similar quality to non-parametric UMAP. To quantitatively measure the quality of embeddings we compared embedding algorithms on several metrics across datasets. We compared each method/dataset on 2D and 64D projections\footnote{Where possible. In contrast with UMAP, Parametric UMAP, and Parametric t-SNE, Barnes Huts t-SNE can only embed in two or three dimensions \citep{van2014accelerating} and while FIt-SNE can in principle scale to higher dimensions \citep{linderman2019fast},  embedding in more than 2 dimensions is unsupported in both the official implementation \citep{fitsne} and openTSNE \citep{Poliar731877}}.
Each metric is explained in detail in \ref{embedding_metrics}. The 2D projection of each dataset/method is shown in Fig \ref{fig:projections}. The results are given in Figs \ref{fig:Trustworthiness}-\ref{fig:nmi} and Tables \ref{table:Trustworthiness}-\ref{table:clustering_nmi}, and summarized below. 

\begin{figure}[!htbp]
  \centering
      \includegraphics[width=1.0\textwidth]{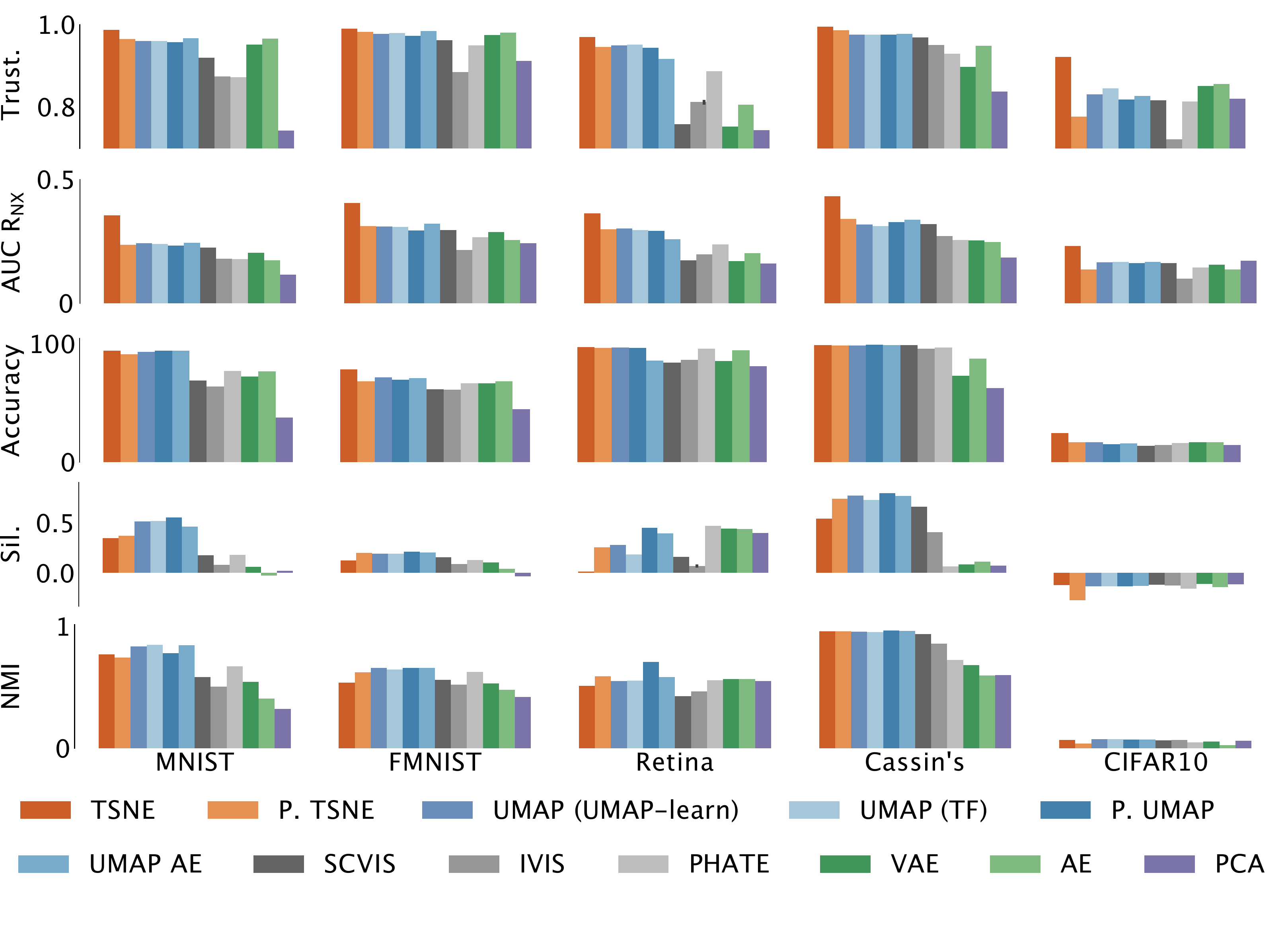}
  \caption{Embedding metrics for 2D projections. Full results are given in the Appendix. Accuracy is shown for KNN (k=1).}
  \label{fig:short_metrics}
\end{figure}

\paragraph{Trustworthiness}
Trustworthiness (Eq. \ref{eq:Trustworthiness}, \citealt{venna2006local}) is a measure of how much of the local structure of a dataset is preserved in a set of embeddings. In 2D, we observe each of the UMAP algorithms performs similarly in Trustworthiness, with t-SNE being slightly more trustworthy in each dataset (Figs \ref{fig:short_metrics},\ref{fig:Trustworthiness}; Table \ref{table:Trustworthiness}). At 64D, PCA, AE, VAE, and Parametric t-SNE are most trustworthy in comparison to each UMAP implementation, possibly reflecting the more approximate repulsion (negative sampling)  used by UMAP. 
\paragraph{Area Under the Curve (AUC) of $R_{NX}$}
\thl{
To compare embeddings across scales (both local and global neighborhoods), we computed the AUC of $R_{NX}$ for each embedding \citep{lee2015multi}, which captures the agreement across K-ary neighborhoods, weighting nearest-neighbors as more important than further neighbors. In 2D we find that Parametric and non-parametric UMAP perform similarly while t-SNE has the highest AUC. At 64D, Parametric and non-parametric UMAP again perform similarly, with PCA having the highest AUC. 
}

\paragraph{KNN-Classifier}
A KNN-classifier is used as a baseline to measure supervised classification accuracy based upon local relationships in embeddings. We find KNN-classifier performance largely reflects Trustworthiness (Figs \ref{fig:short_metrics},\ref{fig:knn1},\ref{fig:knn5}; Tables \ref{table:knn1}, \ref{table:knn5}). In 2D, we observe a broadly similar performance between UMAP and t-SNE variants, each of which is substantially better than the PCA, AE, or VAE projections. At 64 dimensions UMAP projections are similar but in some datasets (FMNIST, CIFAR10) slightly under-performs PCA, AE, VAE, and Parametric t-SNE. 

\paragraph{Silhouette score}
Silhouette score measures how clustered a set of embeddings are given ground truth labels. In 2D, across datasets, we tend to see a better silhouette score for UMAP and Parametric UMAP projections than t-SNE and Parametric t-SNE, which are in turn more clustered than PCA in all cases but CIFAR10, which shows little difference from PCA (Figs \ref{fig:short_metrics},\ref{fig:silhouette}; \ref{table:silhouette}). The clustering of each dataset can also be observed in Fig \ref{fig:projections}, where t-SNE and Parametric t-SNE are more spread out within-cluster than UMAP. In 64D projections, we find the silhouette score of Parametric t-SNE is near or below that of PCA, which is lower than UMAP-based methods. We note, however, that the poor performance of Parametric t-SNE may reflect setting the degrees-of-freedom ($\alpha$) at $d-1$ which is only one of three parameterization schemes that \cite{van2009learning} suggests. A learned degrees-of-freedom parameter might improve performance for parametric t-SNE at higher dimensions.

\paragraph{Clustering}
To compare clustering directly across embeddings, we performed $k$-Means clustering over each latent projection and compared each embedding's clustering on the basis of the normalized mutual information (NMI) between clustering schemes (Figs \ref{fig:short_metrics},\ref{fig:nmi}; Table \ref{table:clustering_nmi}). In both the 2D and 64D projections, we find that NMI corresponds closely to the silhouette score. 
UMAP and t-SNE show comparable clustering in 2D, both well above PCA in most datasets. At 64D, each UMAP approach shows superior performance over t-SNE.

\subsection{Training and embedding speed}

\begin{figure*}[!htbp]
  \centering
      \includegraphics[width=1\textwidth]{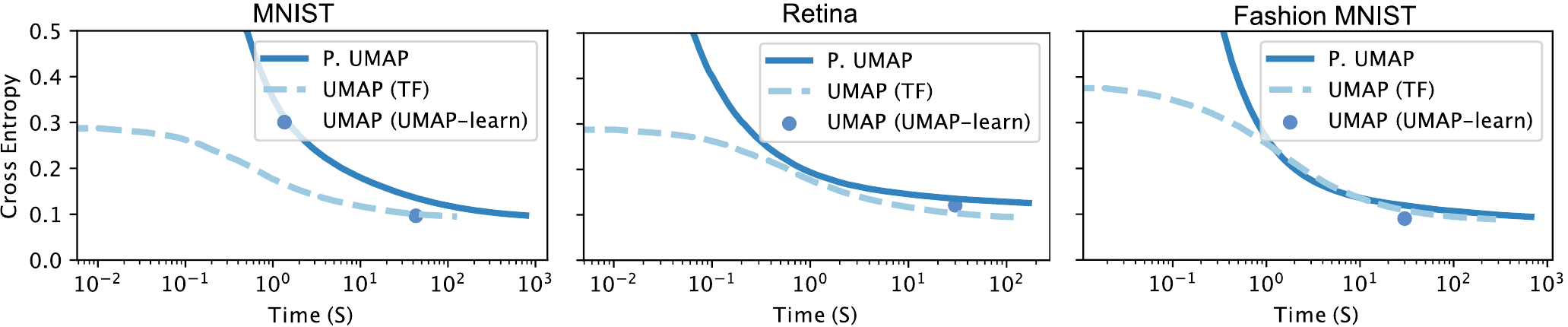}
  \caption{Training times comparison between UMAP and Parametric UMAP. All results were obtained with up to 32 threads on a machine with 2 AMD EPYC Rome 7252 8-Core CPU running at 3.1 GHz and a Quadro RTX 6000.}
  \label{fig:training_time}
\end{figure*}

\paragraph{Training speed} Optimization in non-parametric UMAP is not influenced by the dimensionality of the original dataset; the dimensionality of the dataset only comes into play in computing the nearest-neighbors graph. In contrast, training speeds for Parametric UMAP are variable based upon the dimensionality of data and the architecture of the neural network used. The dimensionality of the embedding does not have a substantial effect on speed. In Fig \ref{fig:training_time}, we show the cross-entropy loss over time for Parametric and non-parametric UMAP, for the MNIST, Fashion MNIST, and Retina datasets. Across each dataset, we find that non-parametric UMAP reaches a lower loss more quickly than Parametric UMAP, but that Parametric UMAP reaches a similar cross-entropy within an order of magnitude of time. Thus, Parametric UMAP can train more slowly than non-parametric UMAP, but training times remain within a similar range making Parametric UMAP a reasonable alternative to non-parametric UMAP in terms of training time.

\begin{figure}[!htbp]
  \centering
      \includegraphics[width=.75\textwidth]{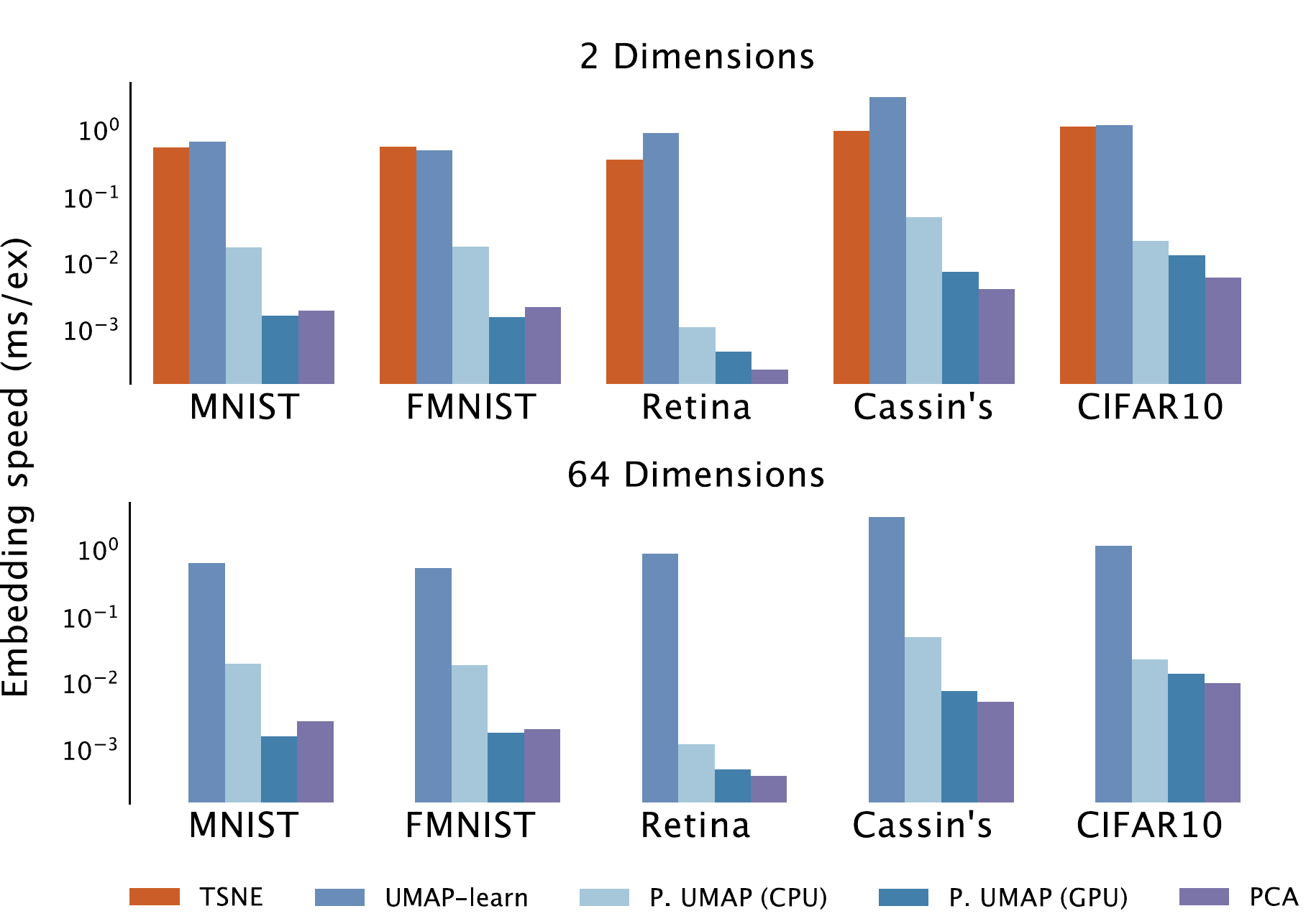}
  \caption{Comparison of embedding speeds using parametric UMAP and other embedding algorithms on a held-out testing dataset. Embeddings were performed on the same machine as Fig \ref{fig:training_time}. Values shown are the median times over 10 runs.}
  \label{fig:embedding_speed}
\end{figure}

\paragraph{Embedding and reconstruction speed} A parametric mapping allows embeddings to be inferred directly from data, resulting in a quicker embedding than non-parametric methods. The speed of embedding is especially important in signal processing paradigms where near-real-time embedding speeds are necessary. For example in brain-machine interfacing, bioacoustics, and computational ethology, fast embedding methods like PCA or deep neural networks are necessary for real-time analyses and manipulations, thus deep neural networks are increasingly being used (e.g. \citealt{pandarinath2018inferring, brown2018ethology, Sainburg870311}). Here, we compare the embedding speed of a held-out test sample for each dataset, as well as the speed of reconstruction of the same held-out test samples.

Broadly, we observe similar embedding times for the non-parametric t-SNE and UMAP methods, which are several orders of magnitude slower than the parametric methods, where embeddings are direct projections into the learned networks (Fig \ref{fig:embedding_speed}). Because the same neural networks are used across the different parametric UMAP and t-SNE methods, we show only Parametric UMAP in Fig 12, which is only slightly slower than PCA, making it a viable candidate for fast embedding where PCA is currently used. Similarly, we compared parametric and non-parametric UMAP reconstruction speeds (Fig \ref{fig:recon_speed}). With the network architectures we used, reconstructions of Parametric UMAP are orders of magnitude faster than non-parametric UMAP, and slightly slower, but within the same order of magnitude, as PCA.

\FloatBarrier

\subsection{Capturing additional global structure in data}
To capture additional global structure we added a na\"{i}ve global structure preservation loss to Parametric UMAP, maximizing the Pearson correlation within batches between pairwise distances in embedding and data spaces:
\begin{equation}
C_{\textrm{\tiny{Pearson}}}=-\frac{\text{cov}(d_X,d_Z)}{\sigma_{d_X} \sigma_{d_Z}}
\label{global_cost}
\end{equation}
Where $\text{cov}(X,Y)$ is the covariance of data and embeddings, and $\sigma_X$ and $\sigma_Z$ are the standard deviations of the data and embeddings.
The same notion of pairwise distance correlation has previously been used directly as a metric for global structure preservation \citep{kobak2021initialization, becht2019dimensionality}.

\begin{figure}[!htbp]
  \centering
      \includegraphics[width=1.0\textwidth]{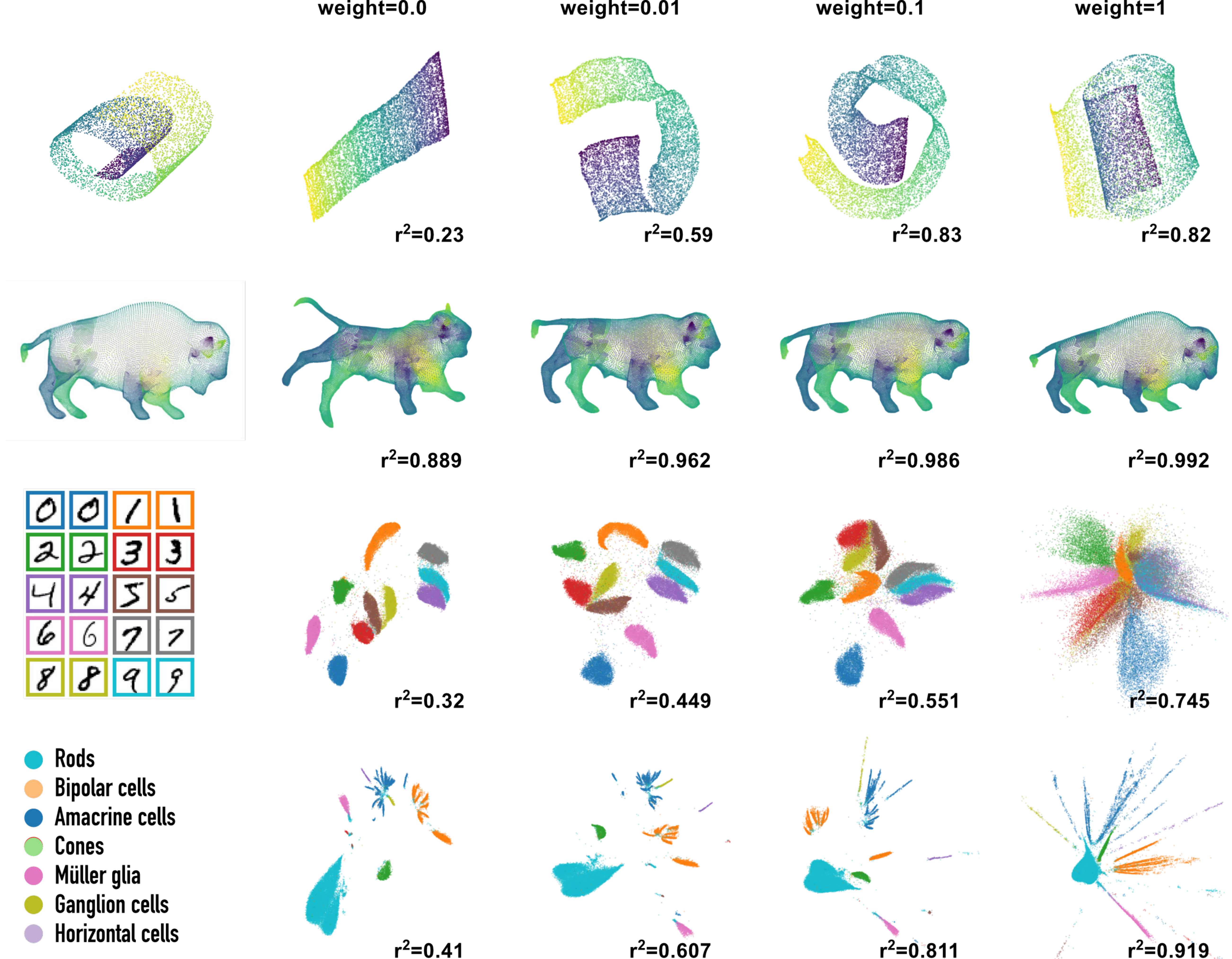}
  \caption{Global loss applied to Parametric UMAP embeddings with different weights. $r^2$ is the correlation between pairwise distances in data space and embedding space.}
  \label{fig:global}
\end{figure}

The weight of this additional loss can be used to dictate the balance between capturing global and local structure in the dataset. In Fig \ref{fig:global}, we apply this loss at four different weights, ranging from only UMAP (left) to primarily global correlation (right). As expected, we observe that as we weight $C_{\textrm{\tiny{Pearson}}}$ more heavily, the global correlation (measured as the correlation of the distance between pairs of points in embedding and data space) increases (indicated in each panel). Notably, when a small weight is used with each dataset, local structure is largely preserved while substantially improving global correlation.

\begin{figure}[!htbp]
  \centering
      \includegraphics[width=1.0\textwidth]{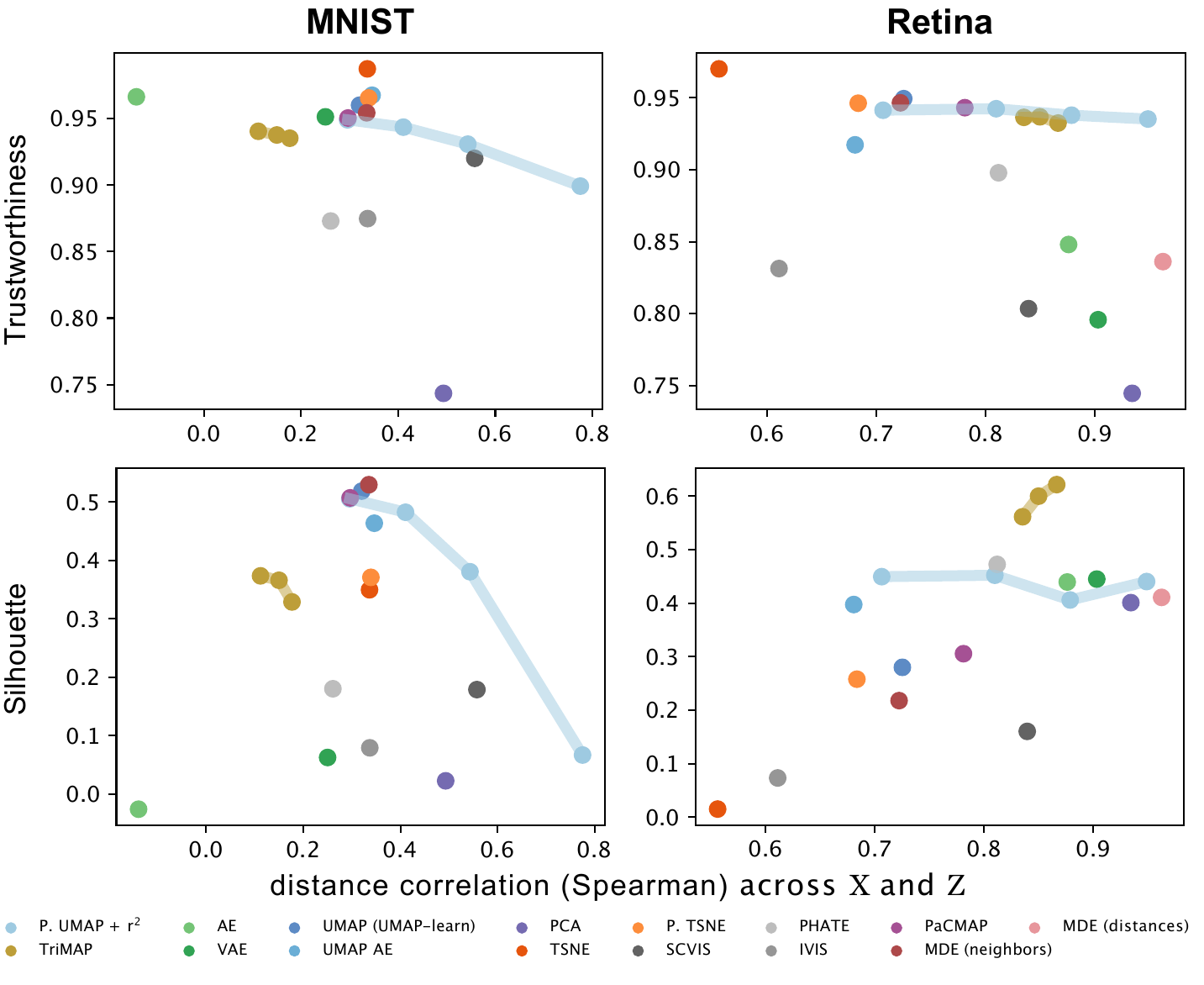}
  \caption{Comparison of pairwise global and local relationship preservation across embeddings for MNIST and Macosko \citeyearpar{macosko2015highly}. Local structure metrics are silhouette score and Trustworthiness. Global structure metric is Spearman correlation of distances in $X$ and $Z$. Connected lines are for the different weights of the correlation loss from Fig \ref{fig:global} in Parametric UMAP, and the $\lambda$ parameter in TriMap (50, 500 and 5000). MDE (distances) is not given for MNIST because of memory issues (on 512Gb RAM).}
  \label{fig:global-vs-local}
\end{figure}

In Fig \ref{fig:global-vs-local}, we show the global distance correlation plotted against two local structure metrics (Silhouette score and Trustworthiness) for the MNIST and Macosko \citeyearpar{macosko2015highly} datasets corresponding to the projections shown in Fig \ref{fig:global} in relation to each embedding from Fig \ref{fig:projections}. In addition, we compared TriMap \citep{amid2019trimap}, a triplet-loss-based embedding algorithm designed to capture additional global structure by preserving the relative distances of triplets of data samples. We also compared Minimum Distortion Embedding (MDE), which comprises two separate embedding functions: a local embedding algorithm that preserves relationships between neighbors similar to UMAP and t-SNE, and a global embedding algorithm that preserves pairwise distances similar to MDS.

Broadly, with Parametric UMAP, we can observe the tradeoff between captured global and local structure with the weight of $C_{\textrm{\tiny{Pearson}}}$ (light blue line in each panel of Fig \ref{fig:global-vs-local}). We observe that adding this loss can increase the amount of global structure captured while preserving much of the local structure, as indicated by the distance to the top right corner of each panel in Fig \ref{fig:global-vs-local}, which reflects the simultaneous capture of global and local relationships, relative to each other embedding algorithm.

\FloatBarrier
\subsection{Autoencoding with UMAP}

The ability to reconstruct data from embeddings can both aid in understanding the structure of non-linear embeddings, as well as allow for manipulation and synthesis of data based on the learned features of the dataset. We compared the reconstruction accuracy across each method which had inverse-transform capabilities (i.e. $Z \rightarrow X$), as well as the reconstruction speed across the neural network-based implementations to non-parametric implementations and PCA. In addition, we performed latent algebra on Parametric UMAP embeddings both with and without an autoencoder regularization and found that reconstructed data can be linearly manipulated in complex feature space.

\begin{figure}[!htbp]
  \centering
      \includegraphics[width=1\textwidth]{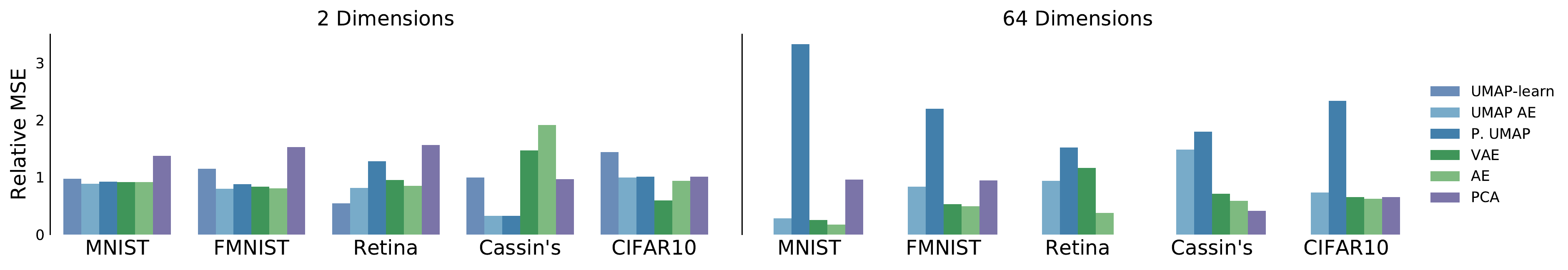}
  \caption{Reconstruction accuracy measured as mean squared error (MSE). MSE is shown relative to each dataset (setting mean at 1).}
  \label{fig:recon_acc}
\end{figure}

\paragraph{Reconstruction accuracy}
We measured reconstruction accuracy as Mean Squared Error (MSE) across each dataset (Fig \ref{fig:recon_acc}; Table \ref{table:recon_results}). In two dimensions, we find that Parametric UMAP typically reconstructs better than non-parametric UMAP, which in turn performs better than PCA. In addition, the autoencoder regularization slightly improves reconstruction performance. At 64 dimensions, the AE regularized Parametric UMAP is generally comparable to the AE and VAE and performs better than Parametric UMAP without autoencoder regularization. The non-parametric UMAP reconstruction algorithm is not compared at 64 dimensions because it relies on an estimation of Delaunay triangulation, which does not scale well with higher dimensions.

\FloatBarrier

\paragraph{Latent features}
Previous work shows that parametric embedding algorithms such as AEs (e.g. Variational Autoencoders) linearize complex data features in latent-space, for example, the presence of a pair of sunglasses in pictures of faces (e.g. \citealt{radford2015unsupervised, white2016sampling, sainburg2018generative}). Here, we performed latent-space algebra and reconstructed manipulations on Parametric UMAP latent-space to explore whether UMAP does the same. 

\begin{figure}[!htbp]
  \centering
     \includegraphics[width=1.0\textwidth]{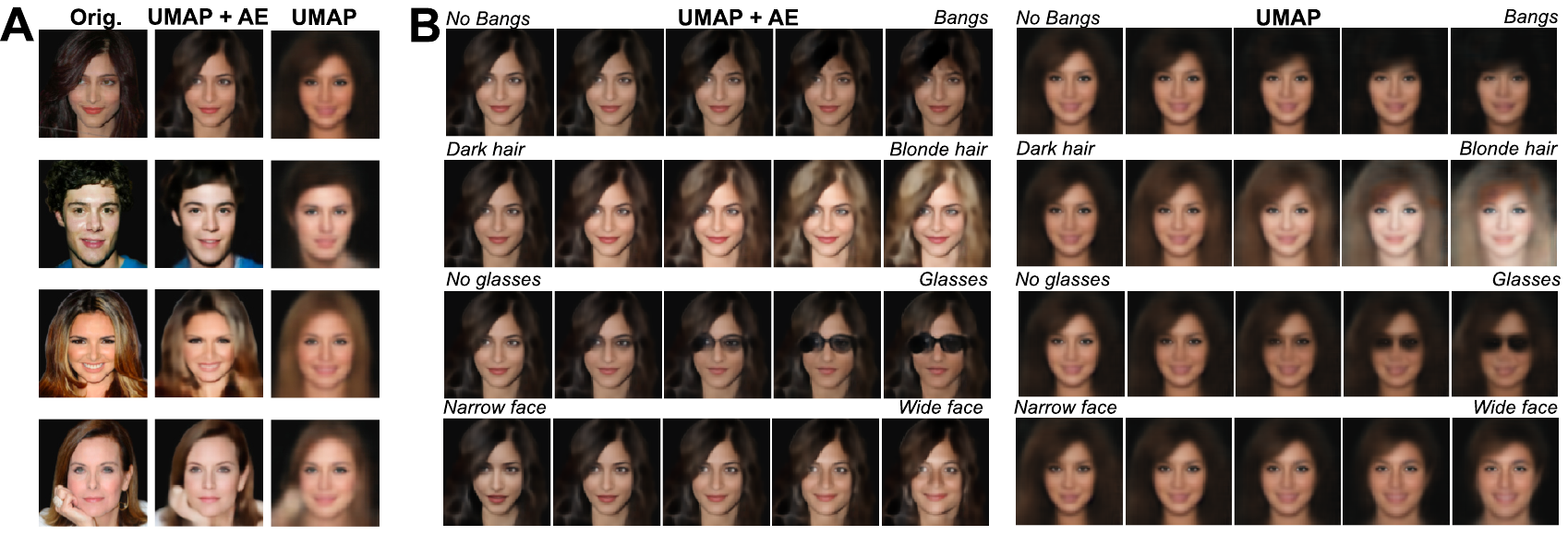}
  \caption{Reconstruction and interpolation. (A) Parametric UMAP reconstructions of faces from a holdout testing dataset. (B) The same networks, adding latent vectors corresponding to image features.}
  \label{fig:latent_features}
\end{figure}

To do so, we use the CelebAMask-HQ dataset, which contains annotations for 40 different facial features over a highly structured dataset of human faces. We projected the dataset of faces into a CNN autoencoder architecture based upon the architecture defined in \cite{huang2018multimodal}. We trained the network first using UMAP loss alone (Parametric UMAP), and second using the joint UMAP and AE loss (Figs \ref{fig:latent_features}). We then fit an OLS regression to predict the latent projections of the entire dataset using the 40 annotated features (e.g. hair color, presence of beard, smiling, etc). The vectors corresponding to each feature learned by the linear model were then treated as feature vectors in latent space and added and subtracted from projected images, then passed through the decoder to observe the resulting image (as in \citealt{sainburg2018generative}). 

We find that complex latent features are linearized in latent space, both when the network is trained with UMAP loss alone as well as when the network is trained with AE loss. For example, in the third set of images in Fig \ref{fig:recon_acc}, a pair of glasses can be added or removed from the projected image by adding or subtracting its corresponding latent vector. 

\subsection{Semi-supervised learning}

Real-word datasets are often comprised of a small number of labeled data, and a large number of unlabeled data. semi-supervised learning (SSL) aims to use the unlabeled data to learn the structure of the dataset, aiding a supervised learning algorithm in making decisions about the data.  
Current SOTA approaches in many areas of supervised learning such as computer vision rely on deep neural networks. Likewise, semi-supervised learning approaches modify supervised networks with structure-learning loss using unlabeled data. 
Parametric UMAP, being a neural network that learns structure from unlabeled data, is well suited to semi-supervised applications. 
Here, we determine the efficacy of UMAP for semi-supervised learning by comparing a neural network jointly trained on classification and UMAP (Fig \ref{fig:networks}D) with a network trained on classification alone using datasets with varying numbers of labeled data. 

We compared datasets ranging from highly-structured (MNIST) to unstructured (CIFAR10) in UMAP using a na\"{i}ve distance metric in data space (e.g. Euclidean distance over images). 
For image datasets, we used a deep convolutional neural network (CNN) which performs with relatively high accuracy for CNN classification on the fully supervised networks (see Table \ref{table:classification_results}) based upon the CNN13 architecture commonly used in SSL \citep{oliver2018realistic}. For the birdsong dataset, we used a BLSTM network, and for the retina dataset, we used a densely connected network. 

\paragraph{Na\"{i}ve UMAP embedding} 

\begin{figure*}[!htbp]
  \centering
    \includegraphics[width=1\textwidth]{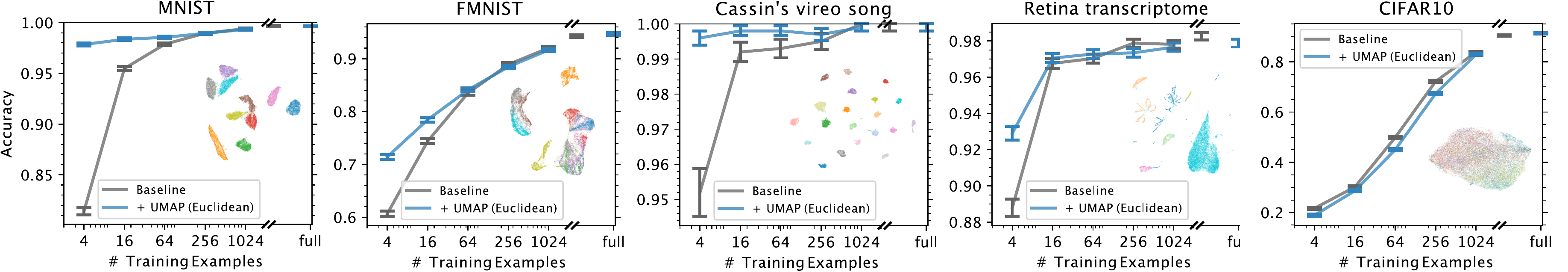}
  \caption{
  Baseline classifier with an additional UMAP loss with different numbers of labeled training examples. Non-parametric UMAP projections of the UMAP graph being jointly trained are shown in the bottom right of each panel. Error bars show SEM.
  }
  \label{fig:ssl_naive}
\end{figure*}

For datasets where structure is learned in UMAP (e.g. MNIST, FMNIST) we expect that regularizing a classifier network with UMAP loss will aid the network in labeling data by learning the structure of the dataset from unlabeled data. To test this, we compared a baseline classifier to a network jointly trained on classifier loss and UMAP loss. We first trained the baseline classifier to asymptotic performance on the validation dataset, then using the pretrained-weights from the baseline classifier, trained a Y-shaped network (Fig \ref{fig:networks}D) jointly on UMAP over Euclidean distances and a classifier loss over the dataset. We find that for each dataset where categorically-relevant structure is found in latent projections of the datasets (MNIST, FMNIST, birdsong, retina), classifications are improved in the semi-supervised network over the supervised network alone, especially with smaller numbers of training examples (Fig \ref{fig:ssl_naive}; Table \ref{table:classification_results}). In contrast, for CIFAR10, the additional UMAP loss impairs performance in the classifier. 

\paragraph{Consistency regularization and learned invariance using data augmentation}

\begin{figure}[!htbp]
  \centering
      \includegraphics[width=0.45\textwidth]{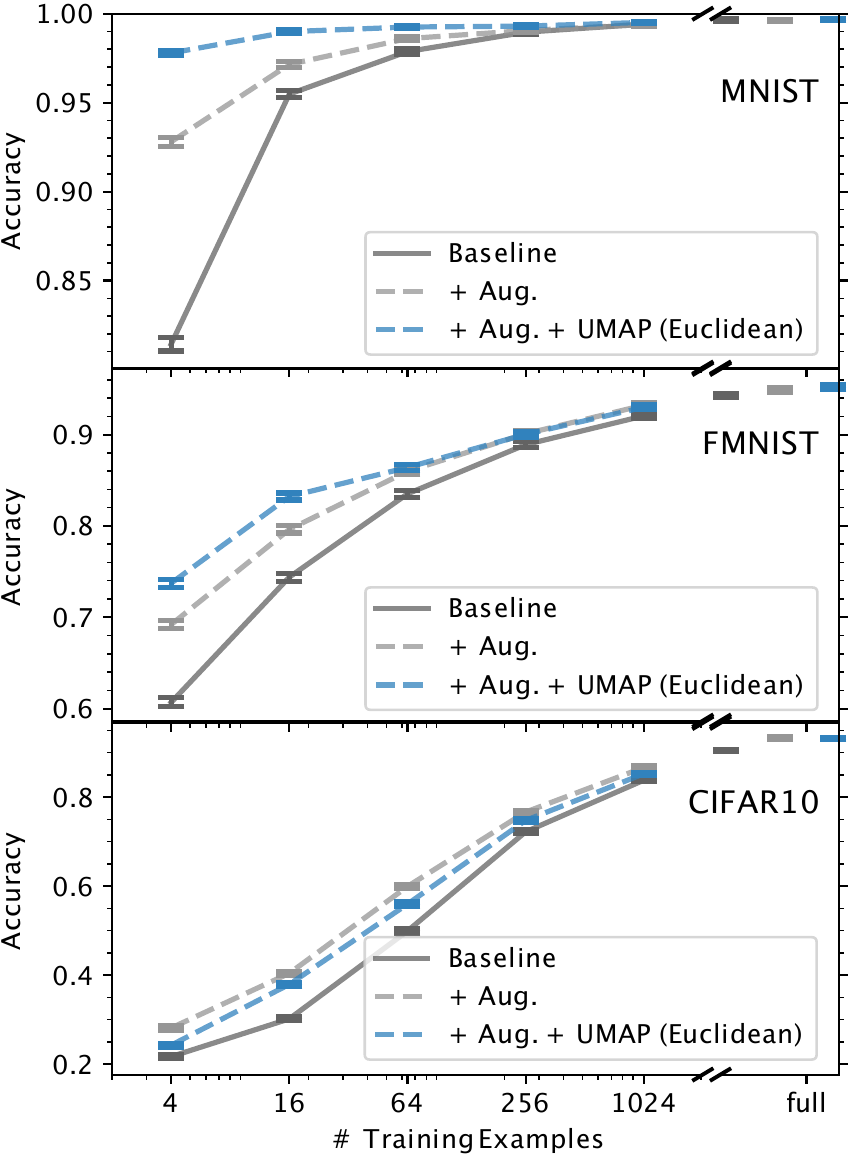}
        \includegraphics[width=0.45\textwidth]{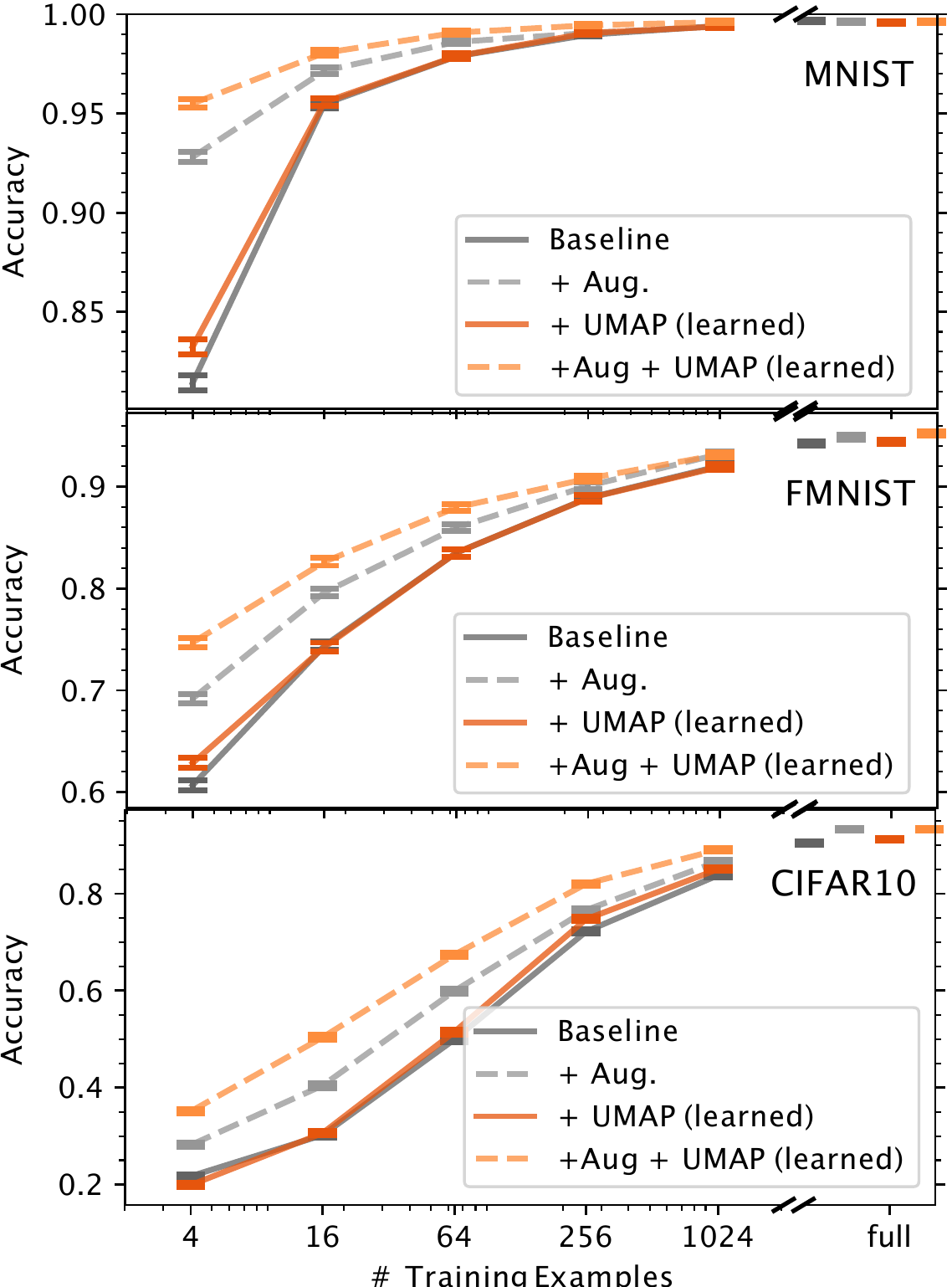}

  \caption{Comparison of baseline classifier, augmentation, and augmentation with an additional UMAP loss (left). SSL using UMAP over the learned latent graph, computed over latent activations in the classifier (right).}
  \label{fig:augmentation}
\end{figure}

Several current SOTA SSL approaches employ a technique called consistency regularization \citep{sajjadi2016regularization}; training a classifier to produce the same predictions with unlabeled data which have been augmented and data that have not been augmented \citep{sohn2020fixmatch, berthelot2019remixmatch}. In a similar vein, for each image dataset, we train the network to preserve the structure of the UMAP graph when data have been augmented. We computed a UMAP graph over un-augmented data and, using augmented data, trained the network jointly using classifier and UMAP loss, teaching the network to learn to optimize the same UMAP graph, invariant to augmentations in the data. We observe a further improvement in network accuracy for MNIST and FMNIST over the baseline, and the augmented baseline (Fig \ref{fig:augmentation} left; Table \ref{table:classification_results}). For the CIFAR10 dataset, the addition of the UMAP loss, even over augmented data, reduces classification accuracy. 

\paragraph{Learning a categorically-relevant UMAP metric using a supervised network}

\begin{figure}[!htbp]
  \centering
      \includegraphics[width=1.0\textwidth]{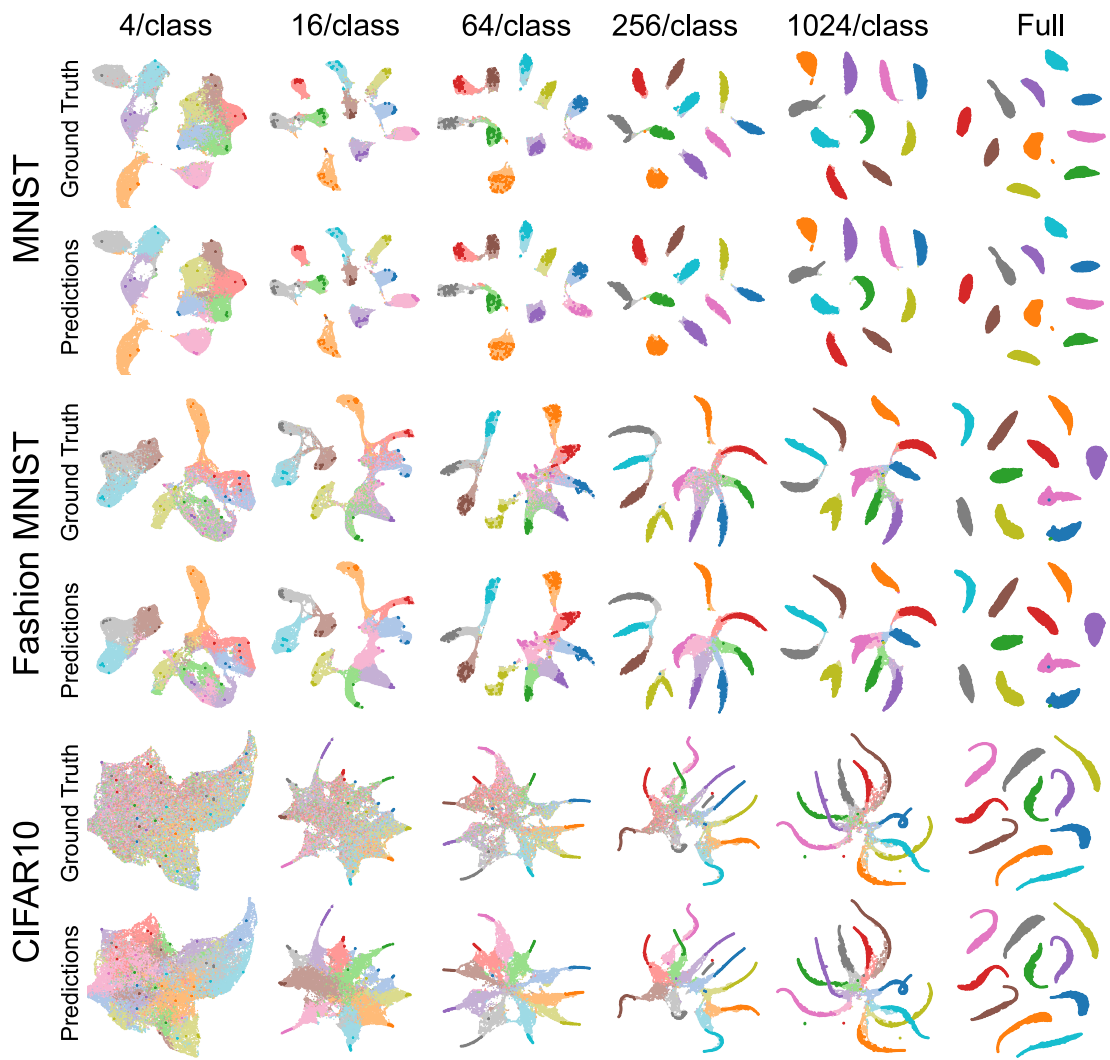}
  \caption{Non-parametric UMAP projections of activations in the last layer of a trained classifier for MNIST, FMNIST, and CIFAR10. For each dataset, the top row shows the ground truth labels on above, and the model's predictions below, in a light colormap. On top of each projection, the labeled datapoints used for training are shown in a darker colormap.}
  \label{fig:latent-classifier-projections}
\end{figure}

It is unsurprising that UMAP confers no improvement for the CIFAR10 dataset, as UMAP computed over the pixel-wise Euclidean distance between images in the CIFAR10 dataset does not capture very much categorically-relevant structure in the dataset. Because no common distance metric over CIFAR10 images is likely to capture such structure, we consider using supervision to learn a categorically-relevant distance metric for UMAP. We do so by training on a UMAP graph computed using distance over latent activations in the classifier network (as in, e.g. \citealt{carter2019activation}), where categorical structure can be seen in UMAP projection (Fig \ref{fig:latent-classifier-projections}). The intuition being that training the network with unlabeled data to capture distributional structure within the network's learned categorically-relevant space will aid in labeling new data.

We find that in all three datasets, without augmentation, the addition of the learned UMAP loss confers little to no improvement in classification accuracy over the data (Fig \ref{fig:augmentation} right; Table \ref{table:classification_results}). When we look at non-parametric projections of the graph over latent activations, we see that the learned graph largely conforms to the network's already-present categorical decision making (e.g. Fig \ref{fig:latent-classifier-projections} predictions vs. ground truth). 
In contrast, with augmentation, the addition of the UMAP loss improves performance in each dataset, including CIFAR10. This contrast in improvement demonstrates that training the network to learn a distribution in a categorically-relevant space that is already intrinsic to the network does not confer any additional information that the network can use in classification. Training the network to be invariant toward augmentations in the data, however, does aid in regularizing the classifier, more in-line with directly training the network on consistency in classifications \citep{sajjadi2016regularization}. 

\subsection{Comparisons with indirect parametric embeddings}
\label{sec:MSE}
In principle, any embedding technique can be implemented parametrically by training a parametric model (e.g. a neural network) to predict embeddings from the original high-dimensional data (as in Duque et al \citeyear{duque2020extendable}). 
However, such a parametric embedding is limited in comparison to directly optimizing the algorithm's loss function. Parametric UMAP optimizes directly over the structure of the graph, with respect to the architecture of the network as well as additional constraints (e.g. additional losses). In contrast, training a neural network to predict non-parametric embeddings does not take additional constraints into account.

\begin{figure}[!htbp]
  \centering
      \includegraphics[width=0.75\textwidth]{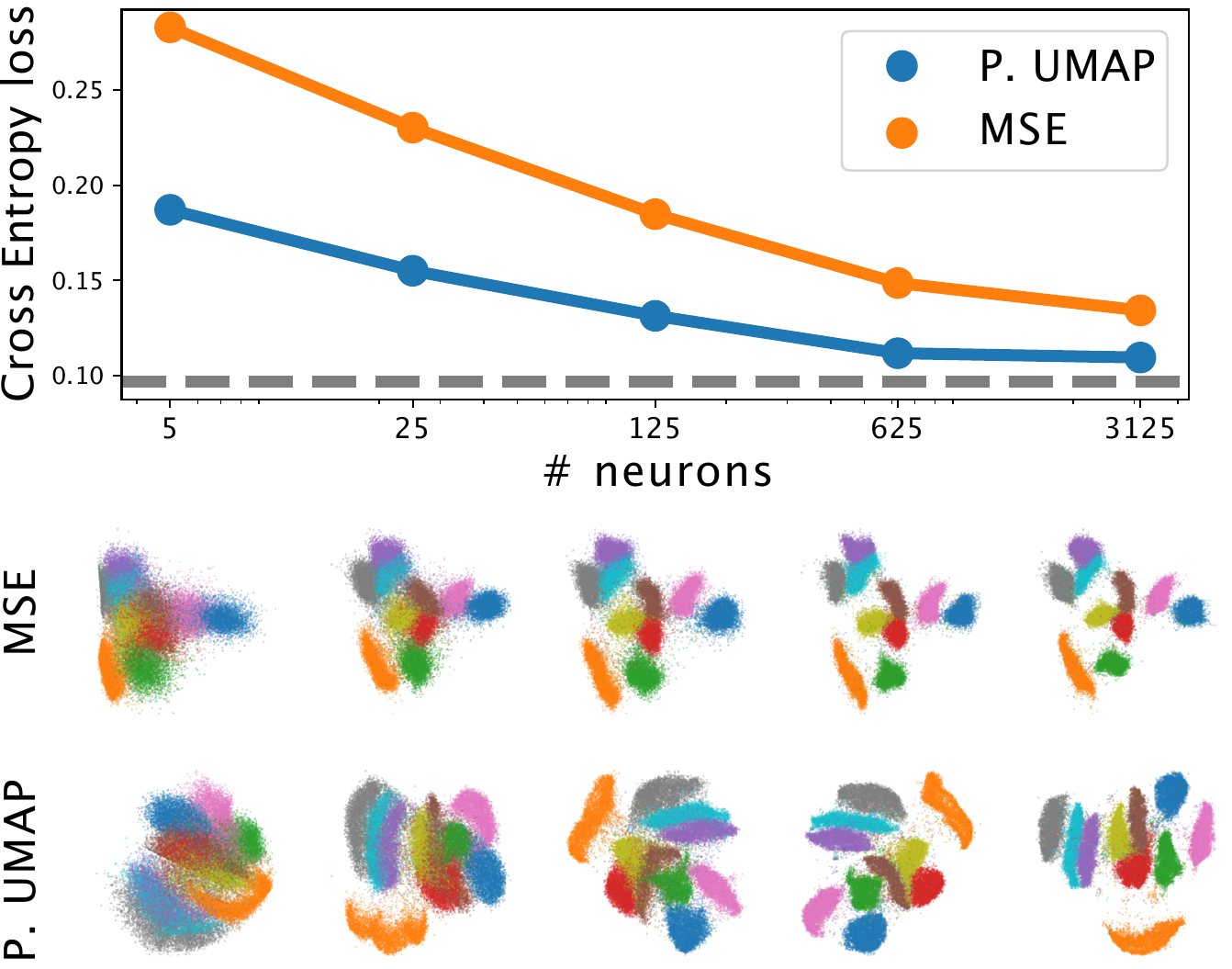}
  \caption{(top) Cross entropy loss for one-hidden-layer instance of Parametric UMAP versus a neural network trained to predict non-parametric embeddings using MSE on MNIST. The same network architectures are used in each case. The$x$-axis varies the number of neurons in the network's single hidden layer layer. The dashed grey line is the loss for the non-parametric embedding. (bottom) Projections corresponding to the losses shown in the panel above.}
  \label{fig:pumap_vs_mse}
\end{figure}

\begin{figure}[!htbp]
  \centering
      \includegraphics[width=1.0\textwidth]{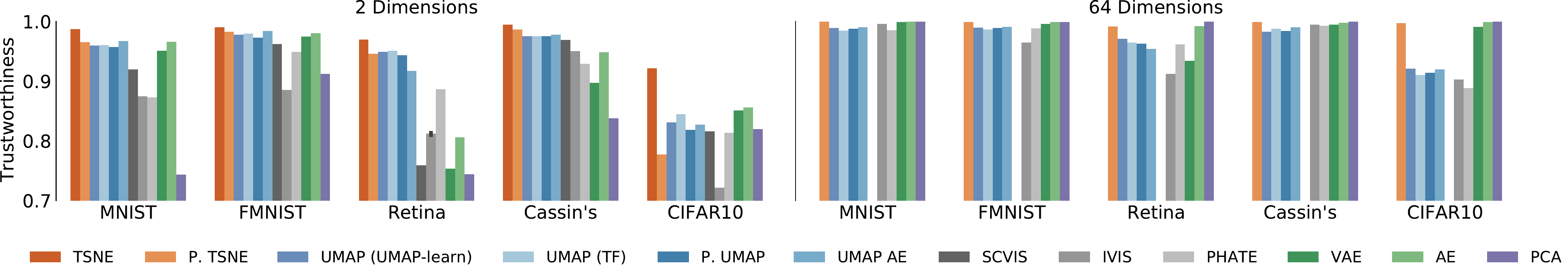}
  \caption{Trustworthiness scores for five datasets using 2- and 64-dimensional projections using each projection method. 64-dimensional t-SNE is not shown due to limitations in high-dimensional projections with t-SNE. Trustworthiness is computed over 10,000 samples of the training dataset. UMAP (TF) stands for the Tensorflow implementation of non-parametric UMAP.}
  \label{fig:Trustworthiness}
\end{figure}

\begin{figure}[!htbp]
  \centering
      \includegraphics[width=1.0\textwidth]{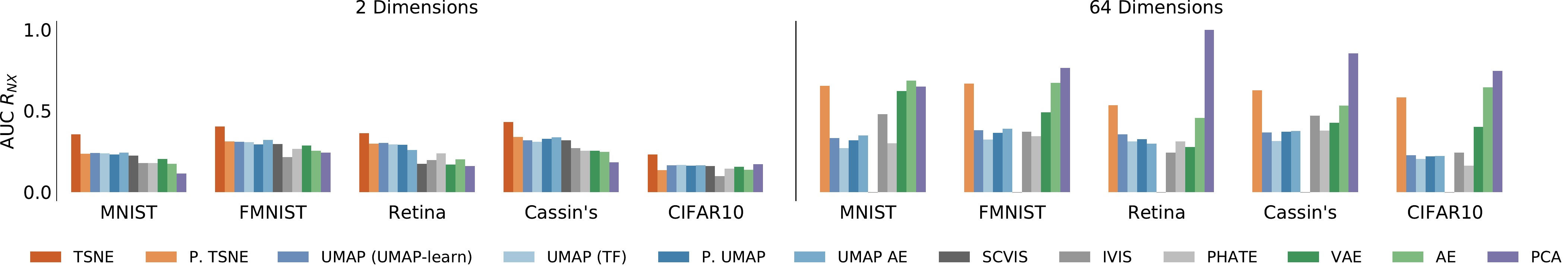}
  \caption{ AUC $R_{NX}$ results on latent projections.}
  \label{fig:R_NX_auc}
\end{figure}

\begin{figure}[!htbp]
  \centering
      \includegraphics[width=1.0\textwidth]{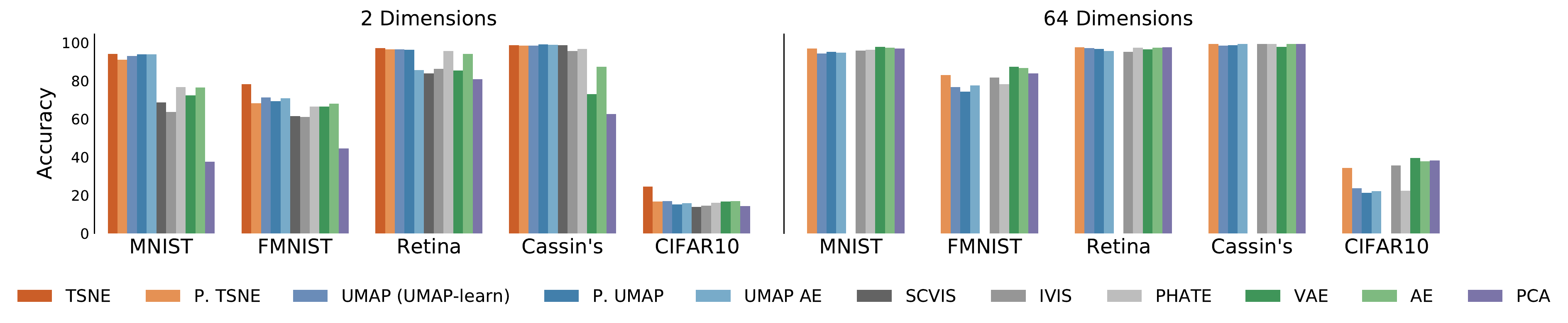}
  \caption{Generalization errors of KNN classifiers (k=1) on latent projections.}
  \label{fig:knn1}
\end{figure}

\begin{figure}[!htbp]
  \centering
      \includegraphics[width=1.0\textwidth]{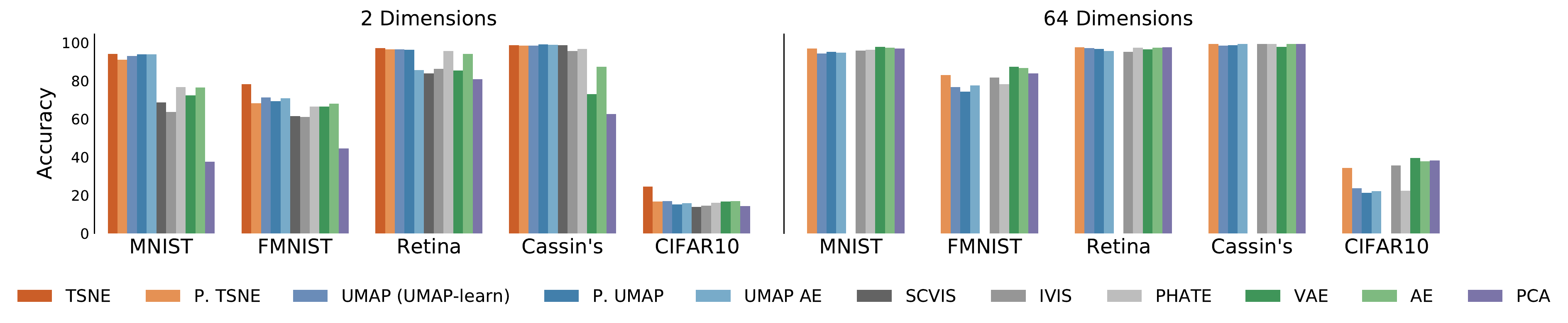}
  \caption{Generalization errors of KNN classifiers (k=5) on latent projections.}
  \label{fig:knn5}
\end{figure}

\begin{figure}[!htbp]
  \centering
      \includegraphics[width=1.0\textwidth]{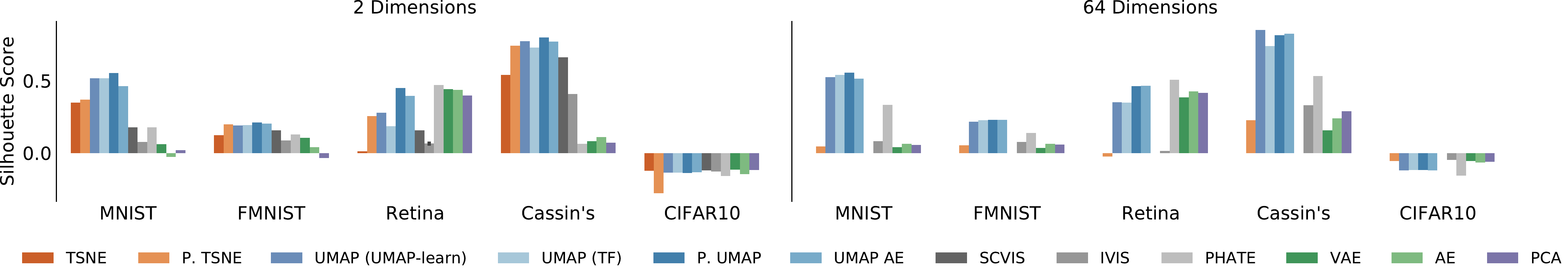}
  \caption{Silhouette scores for five datasets using 2- and 64-dimensional projections using each projection method. 64-dimensional t-SNE is not shown due to limitations in high-dimensional projections with t-SNE.}
\label{fig:silhouette}
\end{figure}

\begin{figure}[!htbp]
  \centering
      \includegraphics[width=1.0\textwidth]{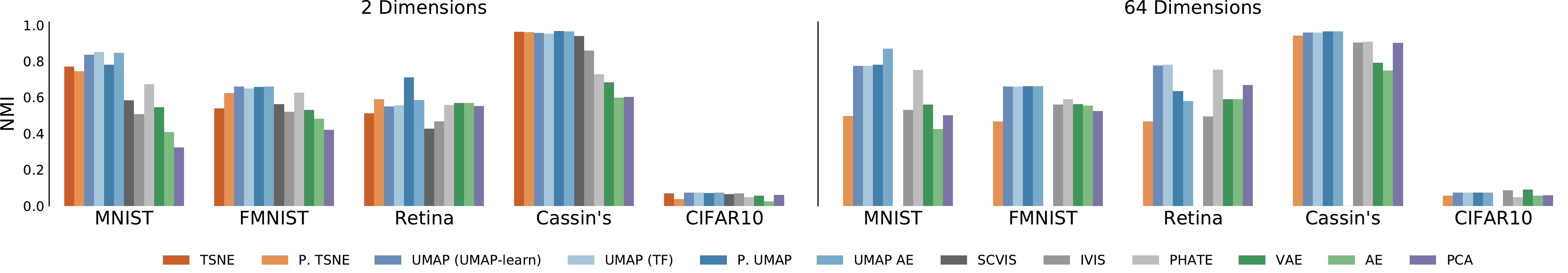}
  \caption{Clustering results. Comparisons are based upon the Normalized Mutual Information (NMI) between labels and clusters. Each dataset shows the NMI for the best clustering chosen on the basis of its silhouette score.}
  \label{fig:nmi}
\end{figure}

\begin{figure}[!htbp]
  \centering
      \includegraphics[width=1\textwidth]{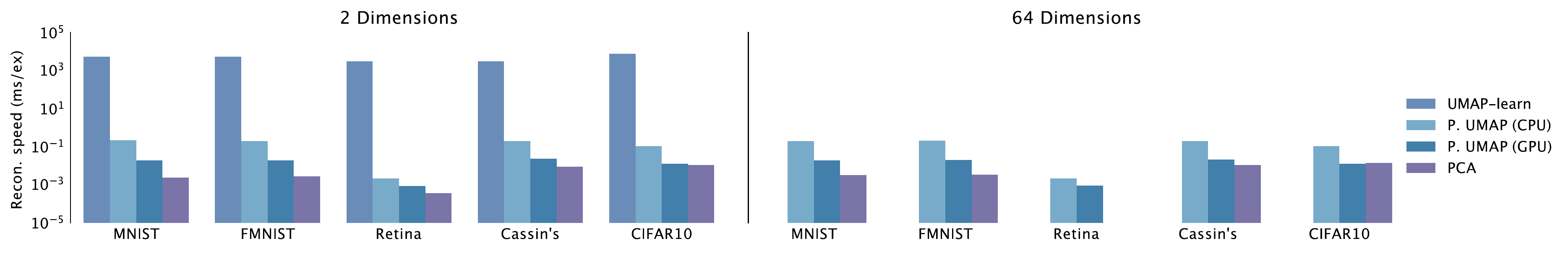}
  \caption{Reconstruction speed. Reconstructions are performed on the same machine as in Fig \ref{fig:embedding_speed}. Reconstructions are not shown for the retina dataset with PCA with a 64D latent space because the dataset is only 50 dimensions. Because all neural network architectures are held constant, speeds remain equal across each Parametric UMAP and t-SNE implementation. Values show the median time over 10 runs.}
  \label{fig:recon_speed}
\end{figure}

To exemplify this, in Fig \ref{fig:pumap_vs_mse} we compare Parametric UMAP to a neural network trained to predict non-parametric embeddings by minimizing MSE when the number of neurons in the network is limited. In the case of Parametric UMAP, the objective of the network is to come up with the best embedding of the UMAP graph that it can, given the constraints of the architecture of the network. In the indirect/MSE case, information about the structure of the graph is only available through an intermediary, the non-parametric embedding, thus this method cannot be optimized to learn an embedding of the data that best preserves the structure of the graph. In other words, the indirect method is not optimizing the embedding of the graph with respect to additional constraints. Instead, it is minimizing the distance between two sets of embeddings. 
The weighted graph is an intermediate topological representation (notably of no specific dimensionality) and is the best representation of the data under UMAP's assumptions. The process of embedding the data in a fixed dimensional space is necessarily a lossy one. Optimizing over the graph directly avoids this loss. This issue also applies when incorporating additional losses (e.g. a classifier loss, or autoencoder loss) to indirect embeddings.

\section{Discussion}

In this paper, we propose a novel parametric extension to UMAP. This parametric form of UMAP produces similar embeddings to non-parametric UMAP, with the added benefit of a learned mapping between data space and embedding space.
We demonstrated the utility of this learned mapping on several downstream tasks. 
We showed that parametric relationships can be used to improve inference times for embeddings and reconstructions by orders of magnitude while maintaining similar embedding quality to non-parametric UMAP. Combined with a global structure preservation loss, Parametric UMAP captures additional global relationships in data, outperforming methods where global structure is only imposed upon initialization (e.g. initializing with PCA embeddings). Combined with an autoencoder, UMAP improves reconstruction quality and allows for the reconstruction of high-dimensional UMAP projections.
We also show that Parametric UMAP projections linearize complex features in latent space.
Parametric UMAP can be used for semi-supervised learning, improving training accuracy on datasets where small numbers of training exemplars are available. We showed that UMAP loss applied to a classifier improves semi-supervised learning in real-world cases where UMAP projections carry categorically-relevant information (such as stereotyped birdsongs or single-cell transcriptomes), but not in cases where categorically-relevant structure is not present (such as CIFAR10). We devised two downstream approaches based around learned categorically-relevant distances, and consistency regularization, that show improvements on these more complex datasets.
Parametric embedding also makes UMAP feasible in fields where dimensionality reduction of continuously generated signals plays an important role in real-time analysis and experimental control.

A number of future directions and extensions to our approach have the potential to further improve upon our results in dimensionality reduction and its various applications. 
\thl{
For example, to improve global structure preservation, we jointly optimized over the Pearson correlation between data and embeddings. Using notions of global structure beyond pairwise distances in data space (such as global UMAP relationships or higher-dimensional simplices) may capture additional structure in data. 
Similarly, one approach we used to improve classifier accuracy relied on obtaining a 'categorically relevant' metric, defined as the Euclidean distance between activation states of the final layer of a classifier. Recent works (e.g. as discussed and proposed in \cite{schulz2019deepview}) have explored methods for more directly capturing class information in the computation of distance, such as using the Fisher metric to capture category- and decision-relevant structure in classifier networks. Similar metrics may prove to further improve semi-supervised classifications with Parametric UMAP.   
}

\section{Acknowledgments}
Work supported by NIH 5T32MH020002-20 to TS and 5R01DC018055-02 to TQG. We would also
like to thank Kyle McDonald for making available his translation of Parametric t-SNE to Tensorflow/Keras, which we used as a basis for our own implementation.

\newpage

\bibliography{example_paper}

\section{Appendix}

\subsection{Datasets}
\label{sec:datasets}
We performed experiments over several different datasets varying in complexity. The \textit{Cassin's vireo song} dataset \citep{hedley2016composition, hedley2016complexity, Sainburg870311} consists of spectrograms of 20 of the most frequently sung elements of Cassin's vireo song (zero-padded to 32 frequency by 31 time bins) produced by several individuals recorded in the Sierra Nevada mountains of California. Despite being recorded in the wild, the Cassin's vireo song is relatively low noise and vocal elements are highly stereotyped. \textit{MNIST} is a benchmark handwritten digits dataset in 28x28 pixels \citep{lecun1998mnist}. \textit{Fashion MNIST} (FMNIST) is a dataset of fashion items in the same format as Fashion MNIST designed to be a more difficult classification problem than MNIST \citep{xiao2017fashion}. \textit{CIFAR10} is a natural image dataset (32x32x3 pixels) with 10 classes \citep{Krizhevsky09learningmultiple}. CIFAR10 classes are much less structured than the other datasets used. For example, unlike FMNIST, subjects of images are not uniformly centered in the image and can have different background conditions that make neighborhood in pixel-space less likely between members of the same class. The \textit{single-cell retina transcriptome} dataset consists of PCA projections (50D) of single-cell RNA transcriptome data from mouse retina \citep{macosko2015highly, Poliar731877}. The \textit{CelebAMask-HQ} dataset consists of cropped and aligned photographs of celebrity faces with 40 facial feature annotations \citep{CelebAMask-HQ, liu2015faceattributes} and label masks corresponding to face-landmarks. We removed the background of each image to make the task of learning a structured embedding simpler for the neural network. A further description of each dataset is given in Table \ref{table:datasets}.

\begin{table*}[!htbp]
\scriptsize
\begin{tabular}{@{}llll@{}}
\toprule
\textbf{Dataset}             & \textbf{Dim.} & \textbf{\# Train/Valid/Test}  & \textbf{Citation} \\ \midrule
\textbf{Moons} &  2         &   1K/NA/NA&      \cite{pedregosa2011scikit}              \\
\textbf{Bison} &  3         &   50K/NA/NA &     \cite{umapzoo}              \\ 
\textbf{Cassin's vireo song} &  32x31         &   24.98K/1K/1K &     \cite{hedley2016composition,hedley2016complexity}              \\
\textbf{MNIST}              &   28x28             &   50K/10K/10K  &   \cite{lecun1998mnist}                \\
\textbf{Fashion MNIST}       &   28x28            &    50K/10K/10K &    \cite{xiao2017fashion}               \\ 
\textbf{CIFAR10}       &      32x32x3         &      40K/10K/10K  &     \cite{Krizhevsky09learningmultiple}              \\ 
\textbf{{Mouse retina  transcriptomes}}       &     50          &       30.33K/4.81K/10K &  \cite{macosko2015highly}                                 \\ 
\textbf{CelebA-HQ (128)}       &   128x128x3            &   28K/1K/1K &     \cite{CelebAMask-HQ, liu2015faceattributes}              \\ \bottomrule
\end{tabular}
\caption{Datasets used across all of the analyses and visualizations in the paper. NA refers to a split that was not used in the paper (e.g. no validation or testing set was used for the bison visualization in Fig \ref{fig:projections}B.)}
\label{table:datasets}
\end{table*}

\subsection{Embedding algorithms}
\label{embedding_algorithms}

Neural network architectures for the Parametric networks differ between datasets. MNIST, FMNIST, and CIFAR10 use convolutional neural networks. The Cassin's vireo dataset uses an LSTM encoder (and decoder for the autoencoder). The Retina dataset uses a 3-layer MLP with 100-neurons per layer. Parametric t-SNE and UMAP used the same neural network architectures and optimizer. For UMAP, the distance metric used for Cassin's vireo song is a dynamic time warping (DTW) metric. The UMAP Autoencoder (UMAP AE) uses the same architecture of the Parametric UMAP and t-SNE implementations, combined with a corresponding decoder network. The encoder network and decoder network are jointly trained on a reconstruction and UMAP loss function. We additionally trained a decoder for the Parametric UMAP network, in which the encoder is trained only on the UMAP loss, and is not jointly trained on a reconstruction loss. For both t-SNE and Parametric t-SNE perplexity was left at its default value of 30 across datasets. We also left the degrees of freedom for parametric t-SNE at $\alpha= d-1$, where $d$ is the number of dimensions in the latent projection. Parametric embeddings (UMAP, t-SNE) are initialized with the same random neural network weights. Non-parametric algorithms use their corresponding default initializations.

\subsection{Embedding metrics}
\label{embedding_metrics}

\subsubsection{Trustworthiness}
\FloatBarrier

Trustworthiness \citep{venna2006local} is a measure of how much of the local structure of a dataset is preserved in a set of embeddings. Trustworthiness is quantified by comparing each datapoint's nearest neighbors in the original space, to its nearest neighbors in the embedding space:

\begin{equation}
\label{eq:Trustworthiness}
T(k)=1-\frac{2}{n k(2 n-3 k-1)} \sum_{i=1}^{n} \sum_{j \in \mathcal{N}_{i}^{k}} \max (0,(r(i, j)-k))
\end{equation}

where $k$ is the number of nearest neighbors Trustworthiness is being computed over, and for each sample $i$, 
$\mathcal{N}_{i}^{k}$ are the nearest $k$ neighbors in the original space, and 
each sample $j$ is $i$'s $r(i, j)$\textsuperscript{th} nearest neighbor.
We compared the Trustworthiness using Scikit-learn's default $k$ value of 5. Trustworthiness is scaled between 0 and 1, with 1 being more trustworthy. 

\paragraph{Area Under the Curve (AUC) of $R_{NX}$}
\thl{
The $R_{NX}(K)$ curve \citep{lee2013type, peluffo2014recent, lee2015multi} measures the (rescaled) agreement between K-ary neighborhoods, and belongs to a class of dimensionality reduction metrics based upon the coranking matrix \citep{lee2009quality}. $R_{NX}(K)$ rescales the average K-ary neighbourhood preservation ($Q_{NX}(K)$) between 0 and 1 such that 1 would be a perfect preservation of neighborhood, and 0 would be no improvement over a random embedding. The AUC of $R_{NX}$ then averages the quality metric $R_{NX}$ across all scales, weighting errors in large neighborhoods proportionally as less important than errors in small neighborhoods (see \cite{lee2015multi} for more detail). 
}

\subsubsection{KNN Classifier}

Related to Trustworthiness, a k-Nearest Neighbor's (KNN) classifier is a supervised algorithm that classifies each unlabeled point based on its k-nearest labeled datapoints. We applied a KNN classifier with k=1 (Fig \ref{fig:knn1}) and k=5 (Fig \ref{fig:knn5}) to each dataset.  

\subsubsection{Silhouette score}
\FloatBarrier
As opposed to Trustworthiness, which measures the preservation of local structure in a projection, the silhouette score \citep{rousseeuw1987silhouettes} measures how 'clustered' a set of embeddings are, given ground truth labels. Silhouette score is computed as the mean silhouette coefficient across embeddings, where the silhouette coefficient is the distance between each embedding and all other embeddings in the same class, minus the distance to the nearest point in a separate class. Silhouette score is scaled between -1 and 1, with 1 being more clustered. 

\subsubsection{Clustering}
To compare clustering directly across embeddings, we performed $k$-Means clustering over each latent projection and compared each embeddings clustering on the basis of the normalized mutual information (NMI) between clustering schemes. For each latent projection, $k$-Means was used to cluster the latent projection with $k$ (the number of clusters) varied between $\frac{1}{2}-1\frac{1}{2}$ times the true number of categories in the dataset. The clustering was repeated five times per $k$. The best clustering was then picked on the basis of the silhouette score between the $k$ clusters and the projections (i.e. without reference to ground truth).

\FloatBarrier

\FloatBarrier
\onecolumn
\section{Figures}

\FloatBarrier
\subsection{Results tables}

\begin{table}[!htbp]
\scriptsize

\begin{tabular}{llrrrrrrrrrrr}
\toprule
     Dataset & Dim.  &    t-SNE &  P. t-SNE &  UMAP &  P. UMAP &  UMAP/AE &   SCVIS &    IVIS &   PHATE &     VAE &  AE &     PCA \\
\midrule
Cassin's & 2  &  \textbf{0.9949} &           0.9867 &      0.9758 &   0.9756 &       0.9777 &  0.9693 &  0.9508 &  0.9292 &  0.8976 &   0.9488 &  0.8380 \\
      & 64 &     - &           0.9990 &      0.9831 &   0.9840 &       0.9907 &     - &  0.9949 &  0.9929 &  0.9949 &   0.9981 &  \textbf{0.9999} \\
CIFAR10 & 2  &  \textbf{0.9216} &           0.7773 &      0.8310 &   0.8187 &       0.8273 &  0.8164 &  0.7221 &  0.8137 &  0.8510 &   0.8564 &  0.8202 \\
      & 64 &     - &           0.9971 &      0.9209 &   0.9140 &       0.9199 &     - &  0.9029 &  0.8885 &  0.9913 &   0.9992 &  \textbf{0.9996} \\
FMNIST & 2  &  \textbf{0.9906} &           0.9827 &      0.9777 &   0.9733 &       0.9842 &  0.9626 &  0.8856 &  0.9494 &  0.9751 &   0.9803 &  0.9126 \\
      & 64 &     - &           0.9991 &      0.9897 &   0.9894 &       0.9913 &     - &  0.9651 &  0.9887 &  0.9960 &   0.9991 &  \textbf{0.9995} \\
Retina & 2  &  \textbf{0.9702} &           0.9463 &      0.9494 &   0.9435 &       0.9173 &  0.7592 &  0.8141 &  0.8871 &  0.7533 &   0.8063 &  0.7445 \\
      & 64 &     - &           0.9918 &      0.9708 &   0.9628 &       0.9542 &     - &  0.9126 &  0.9620 &  0.9342 &   0.9921 &  \textbf{1.0000} \\
MNIST & 2  &  \textbf{0.9874} &           0.9655 &      0.9601 &   0.9573 &       0.9675 &  0.9201 &  0.8748 &  0.8730 &  0.9513 &   0.9663 &  0.7434 \\
      & 64 &     - &           0.9997 &      0.9895 &   0.9880 &       0.9905 &     - &  0.9961 &  0.9856 &  0.9994 &   0.9997 &  \textbf{0.9999} \\
\bottomrule
\end{tabular}

\caption{Trustworthiness score for each method from Fig \ref{fig:Trustworthiness}.}
\label{table:Trustworthiness}
\end{table}

\begin{table}[!htbp]
\scriptsize

\begin{tabular}{llrrrrrrrrrrr}
\toprule
     Dataset & Dim. &    t-SNE &  P. t-SNE &  UMAP &  P. UMAP &  UMAP/AE &   SCVIS &    IVIS &   PHATE &     VAE &  AE &     PCA \\
\midrule
Cassin's & 2  &  \textbf{0.4325} &           0.3405 &      0.3190 &   0.3288 &       0.3375 &  0.3205 &  0.2719 &  0.2563 &  0.2544 &   0.2484 &  0.1848 \\
      & 64 &     - &           0.6280 &      0.3676 &   0.3734 &       0.3767 & -0.0019 &  0.4707 &  0.3788 &  0.4272 &   0.5335 &  \textbf{0.8537} \\
CIFAR10 & 2  &  \textbf{0.2320} &           0.1363 &      0.1664 &   0.1631 &       0.1666 &  0.1619 &  0.1003 &  0.1450 &  0.1559 &   0.1374 &  0.1718 \\
      & 64 &     - &           0.5846 &      0.2281 &   0.2197 &       0.2240 &  0.0002 &  0.2438 &  0.1630 &  0.4022 &   0.6449 &  \textbf{0.7473} \\
FMNIST & 2  &  \textbf{0.4055} &           0.3128 &      0.3106 &   0.2949 &       0.3218 &  0.2967 &  0.2156 &  0.2668 &  0.2883 &   0.2563 &  0.2429 \\
      & 64 &     - &           0.6687 &      0.3807 &   0.3662 &       0.3900 & -0.0004 &  0.3717 &  0.3440 &  0.4922 &   0.6737 &  \textbf{0.7648} \\
Retina & 2  &  \textbf{0.3636} &           0.2985 &      0.3026 &   0.2925 &       0.2590 &  0.1745 &  0.1978 &  0.2381 &  0.1704 &   0.2028 &  0.1604 \\
      & 64 &     - &           0.5350 &      0.3552 &   0.3273 &       0.2987 & -0.0002 &  0.2428 &  0.3130 &  0.2775 &   0.4565 &  \textbf{1.0001} \\
MNIST & 2  &  \textbf{0.3555} &           0.2371 &      0.2422 &   0.2333 &       0.2441 &  0.2244 &  0.1798 &  0.1792 &  0.2041 &   0.1741 &  0.1161 \\
      & 64 &     - &           0.6546 &      0.3334 &   0.3194 &       0.3485 &  0.0010 &  0.4806 &  0.3005 &  0.6230 &   \textbf{0.6867} &  0.6501 \\
\bottomrule
\end{tabular}

\caption{AUC $R_{MX}$ score for each method from Fig \ref{fig:R_NX_auc}
\label{table:R_NX_auc}}
\end{table}

\begin{table}[!htbp]
\scriptsize

\begin{tabular}{llrrrrrrrrrrr}
\toprule
      Dataset & Dim. &    t-SNE &  P. t-SNE &  UMAP &  P. UMAP &  UMAP/AE &   SCVIS &    IVIS &   PHATE &     VAE &  AE &     PCA \\
\midrule
Cassin's & 2  &  0.9880 &           0.9860 &      0.9860 &   \textbf{0.9910} &       0.9890 &  0.9870 &  0.9570 &  0.9690 &  0.7300 &   0.8740 &  0.6260 \\
      & 64 &     - &           \textbf{0.9950} &      0.9850 &   0.9880 &       0.9940 &     - &  0.9950 &  0.9950 &  0.9800 &   0.9950 &  0.9950 \\
CIFAR10 & 2  &  \textbf{0.2457} &           0.1675 &      0.1689 &   0.1512 &       0.1592 &  0.1380 &  0.1445 &  0.1599 &  0.1665 &   0.1696 &  0.1436 \\
      & 64 &     - &           0.3426 &      0.2375 &   0.2139 &       0.2223 &     - &  0.3571 &  0.2249 &  \textbf{0.3949} &   0.3790 &  0.3829 \\
FMNIST & 2  &  \textbf{0.7825} &           0.6834 &      0.7144 &   0.6941 &       0.7083 &  0.6165 &  0.6108 &  0.6655 &  0.6646 &   0.6816 &  0.4467 \\
      & 64 &     - &           0.8300 &      0.7682 &   0.7431 &       0.7772 &     - &  0.8188 &  0.7824 &  \textbf{0.8747} &   0.8671 &  0.8398 \\
Retina & 2  &  \textbf{0.9717} &           0.9661 &      0.9665 &   0.9643 &       0.8581 &  0.8407 &  0.8567 &  0.9583 &  0.8545 &   0.9429 &  0.8085 \\
      & 64 &     - &           \textbf{0.9772} &      0.9721 &   0.9683 &       0.9576 &     - &  0.9538 &  0.9741 &  0.9670 &   0.9750 &  0.9759 \\
MNIST & 2  &  \textbf{0.9411} &           0.9118 &      0.9317 &   0.9402 &       0.9403 &  0.6880 &  0.6369 &  0.7684 &  0.7241 &   0.7647 &  0.3765 \\
      & 64 &     - &           0.9697 &      0.9449 &   0.9518 &       0.9481 &     - &  0.9588 &  0.9634 &  \textbf{0.9785} &   0.9748 &  0.9707 \\
\bottomrule
\end{tabular}

\caption{KNN (k = 1) scores for each method from Fig \ref{fig:knn1}}
\label{table:knn1}
\end{table}

\begin{table}[!htbp]
\scriptsize
\begin{tabular}{llrrrrrrrrrrr}
\toprule
      Dataset & Dim.  &    t-SNE &  P. t-SNE &  UMAP &  P. UMAP &  UMAP/AE &   SCVIS &    IVIS &   PHATE &     VAE &  AE &     PCA \\
\midrule
Cassin's & 2  &  0.9910 &           0.9930 &      0.9890 &   \textbf{0.9950} &       0.9930 &  0.9880 &  0.9740 &  0.9840 &  0.7740 &   0.9090 &  0.6910 \\
      & 64 &     - &           0.9950 &      0.9860 &   0.9910 &       0.9970 &     - &  0.9940 &  \textbf{0.9980} &  0.9880 &   0.9930 &  0.9920 \\
CIFAR10 & 2  &  \textbf{0.2608} &           0.2017 &      0.1936 &   0.1722 &       0.1833 &  0.1584 &  0.1592 &  0.1815 &  0.1941 &   0.2007 &  0.1503 \\
      & 64 &     - &           0.3556 &      0.2694 &   0.2519 &       0.2477 &     - &  \textbf{0.3800} &  0.2517 &  0.3777 &   0.3728 &  0.3769 \\
FMNIST & 2  &  \textbf{0.8039} &           0.7361 &      0.7608 &   0.7407 &       0.7561 &  0.6735 &  0.6620 &  0.7138 &  0.7161 &   0.7339 &  0.5055 \\
      & 64 &     - &           0.8479 &      0.8059 &   0.7878 &       0.8028 &     - &  0.8355 &  0.8126 &  \textbf{0.8830} &   0.8756 &  0.8568 \\
Retina & 2  &  \textbf{0.9795} &           0.9766 &      0.9792 &   0.9761 &       0.8933 &  0.8864 &  0.8942 &  0.9679 &  0.8795 &   0.9647 &  0.8429 \\
      & 64 &     - &           0.9813 &      0.9801 &   0.9748 &       0.9661 &     - &  0.9654 &  0.9790 &  0.9770 &   \textbf{0.9817} &  0.9806 \\
MNIST & 2  &  0.9502 &           0.9378 &      0.9544 &   \textbf{0.9614} &       0.9537 &  0.7355 &  0.7009 &  0.8114 &  0.7649 &   0.7926 &  0.4201 \\
      & 64 &     - &           0.9734 &      0.9538 &   0.9680 &       0.9654 &     - &  0.9614 &  0.9680 &  \textbf{0.9791} &   0.9758 &  0.9727 \\
\bottomrule
\end{tabular}
\caption{KNN (k = 5) scores for each method from Fig \ref{fig:knn1}}
\label{table:knn5}
\end{table}

\begin{table}[!htbp]
\scriptsize
\begin{tabular}{llrrrrrrrrrrr}
\toprule
      Dataset & Dim. &    t-SNE &  P. t-SNE &  UMAP &  P. UMAP &  UMAP/AE &   SCVIS &    IVIS &   PHATE &     VAE &  AE &     PCA \\
\midrule
Cassin's & 2  &  0.5431 &           0.7439 &      0.7749 &   \textbf{0.8013} &       0.7714 &  0.6643 &  0.4093 &  0.0650 &  0.0853 &   0.1125 &  0.0731 \\
      & 64 &     - &           0.2299 &      \textbf{0.8536} &   0.8173 &       0.8271 &     - &  0.3332 &  0.5333 &  0.1583 &   0.2411 &  0.2914 \\
CIFAR10 & 2  & -0.1216 &          -0.2757 &     -0.1340 &  -0.1359 &      -0.1320 & -0.1190 & -0.1261 & -0.1573 & \textbf{-0.1114} &  -0.1436 & -0.1142 \\
      & 64 &     - &          -0.0536 &     -0.1166 &  -0.1163 &      -0.1172 &     - & \textbf{-0.0445} & -0.1550 & -0.0529 &  -0.0644 & -0.0580 \\
FMNIST & 2  &  0.1251 &           0.2013 &      0.1936 &   \textbf{0.2139} &       0.2060 &  0.1581 &  0.0882 &  0.1298 &  0.1064 &   0.0427 & -0.0331 \\
      & 64 &     - &           0.0543 &      0.2195 &   \textbf{0.2315} &       0.2305 &     - &  0.0790 &  0.1405 &  0.0376 &   0.0655 &  0.0618 \\
Retina & 2  &  0.0151 &           0.2578 &      0.2800 &   0.4519 &       0.3973 &  0.1603 &  0.0659 &  \textbf{0.4723} &  0.4449 &   0.4394 &  0.4009 \\
      & 64 &     - &          -0.0214 &      0.3522 &   0.4652 &       0.4662 &     - &  0.0178 &  \textbf{0.5073} &  0.3873 &   0.4289 &  0.4188 \\
MNIST & 2  &  0.3498 &           0.3710 &      0.5186 &   \textbf{0.5559} &       0.4637 &  0.1789 &  0.0793 &  0.1803 &  0.0627 &  -0.0258 &  0.0228 \\
      & 64 &     - &           0.0488 &      0.5276 &   \textbf{0.5571} &       0.5166 &     - &  0.0842 &  0.3342 &  0.0431 &   0.0653 &  0.0569 \\
\bottomrule
\end{tabular}

\caption{Silhouette score for each method from Fig \ref{fig:silhouette}.}
\label{table:silhouette}
\end{table}

\begin{table}[!htbp]
\scriptsize

\begin{tabular}{llrrrrrrrrrrr}
\toprule
     Dataset  & Dim.  &    t-SNE &  P. t-SNE &  UMAP &  P. UMAP &  UMAP/AE &   SCVIS &    IVIS &   PHATE &     VAE &  AE &     PCA \\
\midrule
Cassin's & 2  &  0.9628 &           0.9605 &      0.9581 &   \textbf{0.9686} &       0.9660 &  0.9399 &  0.8602 &  0.7275 &  0.6836 &   0.5984 &  0.6029 \\
      & 64 &     - &           0.9434 &      0.9596 &   \textbf{0.9665} &       0.9662 &     - &  0.9043 &  0.9094 &  0.7923 &   0.7501 &  0.9019 \\
CIFAR10 & 2  &  0.0688 &           0.0383 &      \textbf{0.0743} &   0.0719 &       0.0730 &  0.0650 &  0.0699 &  0.0485 &  0.0560 &   0.0258 &  0.0605 \\
      & 64 &     - &           0.0570 &      0.0733 &   0.0742 &       0.0746 &     - &  0.0859 &  0.0485 &  \textbf{0.0905} &   0.0574 &  0.0599 \\
FMNIST & 2  &  0.5408 &           0.6248 &      \textbf{0.6603} &   0.6594 &       0.6602 &  0.5627 &  0.5217 &  0.6266 &  0.5319 &   0.4818 &  0.4221 \\
      & 64 &     - &           0.4680 &      0.6602 &   0.6618 &       \textbf{0.6635} &     - &  0.5612 &  0.5905 &  0.5639 &   0.5541 &  0.5244 \\
Retina & 2  &  0.5124 &           0.5912 &      0.5510 &   \textbf{0.7112} &       0.5862 &  0.4278 &  0.4671 &  0.5587 &  0.5691 &   0.5695 &  0.5522 \\
      & 64 &     - &           0.4682 &      \textbf{0.7763} &   0.6356 &       0.5793 &     - &  0.4951 &  0.7539 &  0.5908 &   0.5897 &  0.6696 \\
MNIST & 2  &  0.7704 &           0.7446 &      0.8375 &   0.7824 &       \textbf{0.8460} &  0.5841 &  0.5082 &  0.6728 &  0.5467 &   0.4093 &  0.3234 \\
      & 64 &     - &           0.4977 &      0.7747 &   0.7818 &       \textbf{0.8701} &     - &  0.5322 &  0.7522 &  0.5617 &   0.4253 &  0.5010 \\
\bottomrule
\end{tabular}

\caption{Clustering score for each method from Fig \ref{fig:nmi}}
\label{table:clustering_nmi}
\end{table}

\begin{table}[!htbp]
\scriptsize
\begin{tabular}{llrrrrrr}
\toprule
Dataset & Dim. &   UMAP &  P. UMAP &      UMAP/AE &  AE &     VAE &     PCA \\
\midrule
Cassin's & 2  &      0.0085 &   \textbf{0.0028} &  \textbf{0.0028} &   0.0163 &  0.0125 &  0.0082 \\
      & 64 &         - &   0.0034 &  0.0028 &   0.0011 &  0.0013 &  \textbf{0.0008} \\
CIFAR10 & 2  &      0.0528 &   0.0369 &  0.0364 &   0.0344 &  \textbf{0.0217} &  0.0370 \\
      & 64 &         - &   0.0300 &  0.0094 &   \textbf{0.0080} &  0.0084 &  0.0084 \\
FMNIST & 2  &      0.0347 &   0.0266 &  \textbf{0.0240} &   0.0244 &  0.0253 &  0.0461 \\
      & 64 &         - &   0.0241 &  \textbf{0.0092} &   0.0054 &  0.0058 &  0.0104 \\
Retina & 2  &      \textbf{0.0003} &   0.0008 &  0.0005 &   0.0005 &  0.0006 &  0.0010 \\
      & 64 &         - &   0.0005 &  0.0003 &   \textbf{0.0001} &  0.0003 &     - \\
MNIST & 2  &      0.0393 &   0.0374 &  \textbf{0.0360} &   0.0369 &  0.0371 &  0.0557 \\
      & 64 &         - &   0.0313 &  0.0027 &   \textbf{0.0016} &  0.0024 &  0.0090 \\
\bottomrule
\end{tabular}
\caption{Reconstruction error on held-out testing set for each method.}
\label{table:recon_results}
\end{table}

\begin{table}[!htbp]
\scriptsize
\begin{tabular}{llrrrrr}
\toprule
            &                &      4 &     64 &    256 &   1024 &   full \\
\midrule
MNIST & Baseline &  0.814 &  0.979 &  0.990 &  0.994 &  0.996 \\
            & + Aug. &  0.928 &  0.986 &  0.990 &  0.994 &  0.996 \\
            & + UMAP (Euclidean) &  \textbf{0.978} &  0.986 &  0.990 &  0.993 &  0.996 \\
            & + UMAP (learned) &  0.832 &  0.979 &  0.990 &  0.994 &  0.996 \\
            & + Aug. + UMAP (learned) &  0.955 &  0.991 &  \textbf{0.994} &  \textbf{0.996} &  0.996 \\
            & + Aug. + UMAP (Euclidean) &  \textbf{0.978} &  \textbf{0.992} &  0.993 &  0.995 &  \textbf{0.997} \\
\midrule
FMNIST & Baseline &  0.607 &  0.835 &  0.889 &  0.920 &  0.943 \\
            & + Aug. &  0.692 &  0.860 &  0.901 &  \textbf{0.932} &  0.949 \\
            & + UMAP (Euclidean) &  0.714 &  0.841 &  0.885 &  0.916 &  0.947 \\
            & + UMAP (learned) &  0.629 &  0.835 &  0.889 &  0.920 &  0.944 \\
            & + Aug. + UMAP (learned) &  \textbf{0.747} &  \textbf{0.880} &  \textbf{0.908} &  \textbf{0.932} &  \textbf{0.952} \\
            & + Aug. + UMAP (Euclidean) &  0.737 &  0.864 &  0.900 &  0.930 &  \textbf{0.952} \\
\midrule
CIFAR10 & Baseline &  0.217 &  0.499 &  0.722 &  0.838 &  0.905 \\
            & + Aug. &  0.281 &  0.599 &  0.766 &  0.867 &  0.933 \\
            & + UMAP (Euclidean) &  0.190 &  0.450 &  0.674 &  0.829 &  0.913 \\
            & + UMAP (learned) &  0.199 &  0.515 &  0.748 &  0.850 &  0.912 \\
            & + Aug. + UMAP (learned) &  \textbf{0.351} &  \textbf{0.674} &  \textbf{0.820} &  \textbf{0.891} &  0\textbf{.932} \\
            & + Aug. + UMAP (Euclidean) &  0.243 &  0.560 &  0.748 &  0.852 &  \textbf{0.932} \\
\midrule
Cassin's vireo & Baseline &  0.952 &  0.993 &  0.995 &  \textbf{1.000} &  \textbf{0.999} \\
            & + UMAP (Euclidean) &  \textbf{0.996} &  \textbf{0.998} & \textbf{ 0.997} &  0.999 &  \textbf{0.999} \\
\midrule
Retina & Baseline &  0.888 &  0.970 &  \textbf{0.979} &  \textbf{0.978} &  \textbf{0.983} \\
            & + UMAP (Euclidean) &  \textbf{0.929} &  \textbf{0.973} &  0.973 &  0.977 &  0.979 \\
\bottomrule
\end{tabular}

\caption{Classification accuracy across each dataset and method for different numbers of labeled training examples.}
\label{table:classification_results}

\end{table}

\end{document}